\documentclass{article}

\PassOptionsToPackage{numbers,sort,compress}{natbib}
\usepackage[nonatbib,preprint]{neurips_2020}

\usepackage[utf8]{inputenc} % allow utf-8 input
\usepackage[T1]{fontenc}    % use 8-bit T1 fonts
\usepackage{hyperref}       % hyperlinks
\usepackage{url}            % simple URL typesetting
\usepackage{booktabs}       % professional-quality tables
\usepackage{amsfonts}       % blackboard math symbols
\usepackage{nicefrac}       % compact symbols for 1/2, etc.
\usepackage{microtype}      % microtypography
\usepackage{graphicx}
\usepackage{amssymb}
\usepackage{makecell}
\usepackage{multirow}
\usepackage{wrapfig}
\usepackage{xr}
\usepackage[title]{appendix}
\usepackage{minitoc}

\usepackage{adjustbox}
\usepackage{subcaption}

\usepackage[table]{xcolor}

\usepackage{amsmath}
\usepackage{enumitem}

\DeclareFontFamily{U}{wncy}{}
\DeclareFontShape{U}{wncy}{m}{n}{<->wncyr10}{}
\DeclareSymbolFont{mcy}{U}{wncy}{m}{n}
\DeclareMathSymbol{\Sh}{\mathord}{mcy}{"58} 

\definecolor{yellow}{rgb}{0.6, 0.6, 0.2}
\definecolor{darkyellow}{rgb}{0.8, 0.8, 0.5}
\definecolor{orange}{rgb}{1, 0.8, 0.6}
\definecolor{red}{rgb}{1, 0.3, 0.3}
\definecolor{darkred}{rgb}{0.7, 0.3, 0.3}
\definecolor{darkgreen}{rgb}{0.3, 0.7, 0.3}
\definecolor{blue}{rgb}{0, 0, 1.0}
\definecolor{green}{rgb}{0, 1.0, 0}
\definecolor{pink}{rgb}{1, 0.4, 0.7}

\newcommand{\norm}[1]{\left\lVert#1\right\rVert}

\newcommand{\R}{\mathbb{R}}  % Real numbers
\newcommand{\etal}{\textit{et al}. }
\newcommand\etc{etc\@ifnextchar.{}{.\@}}
\newcommand{\ie}{\textit{i}.\textit{e}., }
\newcommand{\eg}{\textit{e}.\textit{g}. }

\newcommand{\hNTK}{h_{\mathrm{NTK}}}
\newcommand{\loss}{\mathcal{L}}  % Loss function
\newcommand{\D}{\mathcal{D}}  % Data distribution
\newcommand{\N}{\mathcal N}  % Normal distribution

\newcommand{\transpose}{\mathrm{T}}
\newcommand{\testvaleq}{3}

\newcommand{\val}{{\mathrm{val}}}
\newcommand{\opt}{{\mathrm{opt}}}
\newcommand{\lin}{\mathrm{lin}}

\newcommand{\mat}[1]{\mathbf{#1}}
\newcommand{\arbitrary}[1]{#1_{\mathrm{test}}}

\newcommand{\x}{\mathbf{x}}

\newcommand{\uvec}{\mathbf{u}}
\newcommand{\vvec}{\mathbf{v}}
\newcommand{\ltwo}{L2~}
\newcommand{\Data}{\mat{X}}
\newcommand{\Linear}{\mat{A}}
\newcommand{\Kernel}{\mat{K}}

\newcommand{\Eigvec}{\mat{Q}}
\newcommand{\Eigval}{\mat{\Lambda}}
\newcommand{\Eye}{\mat{I}}
\newcommand{\num}{n}

\newcommand{\X}{\Data}  % Training data
\newcommand{\Y}{\mathbf{y}}  % Training labels

\newcommand{\Vector}[1]{\left[#1\right]^\transpose}

\newcommand{\shortpara}[1]{{\bf #1}}

\makeatletter

\makeatletter
\newcommand{\printfnsymbol}[1]{%
        \textsuperscript{\@fnsymbol{#1}}%
}
\makeatother

\title{
Fourier Features Let Networks Learn \\ High Frequency Functions in Low Dimensional Domains
}

\author{
    Matthew Tancik$^{1}$\printfnsymbol{1}
    \And
    Pratul P. Srinivasan$^{1,2}$\printfnsymbol{1}
    \And
    Ben Mildenhall$^1$\printfnsymbol{1}
    \And
    Sara Fridovich-Keil$^1$
    \AND  
    Nithin Raghavan$^1$
    \And 
    Utkarsh Singhal$^1$
    \And 
    Ravi Ramamoorthi$^3$
    \And 
    Jonathan T. Barron$^2$
    \And
    Ren Ng$^1$
    \And
    \vspace{-0.6cm} \\
    $^1$University of California, Berkeley \qquad  $^2$Google Research \qquad $^3$University of California, San Diego \\
}

\begin{document}
\doparttoc % Tell to minitoc to generate a toc for the parts
\faketableofcontents

\maketitle

\begin{abstract}
We show that passing input points through a simple Fourier feature mapping enables a multilayer perceptron (MLP) to learn high-frequency functions in low-dimensional problem domains. These results shed light on recent advances in computer vision and graphics that achieve state-of-the-art results by using MLPs to represent complex 3D objects and scenes. Using tools from the neural tangent kernel (NTK) literature, we show that a standard MLP fails to learn high frequencies both in theory and in practice. To overcome this spectral bias, we use a Fourier feature mapping to transform the effective NTK into a stationary kernel with a tunable bandwidth. We suggest an approach for selecting problem-specific Fourier features that greatly improves the performance of MLPs for low-dimensional regression tasks relevant to the computer vision and graphics communities.
\end{abstract}

\section{Introduction}

\begin{figure}[t]
\centering
\begin{tabular}{@{}c@{\,\,}c@{\,\,}c@{}}
\makecell{
\includegraphics[width=0.175\textwidth]{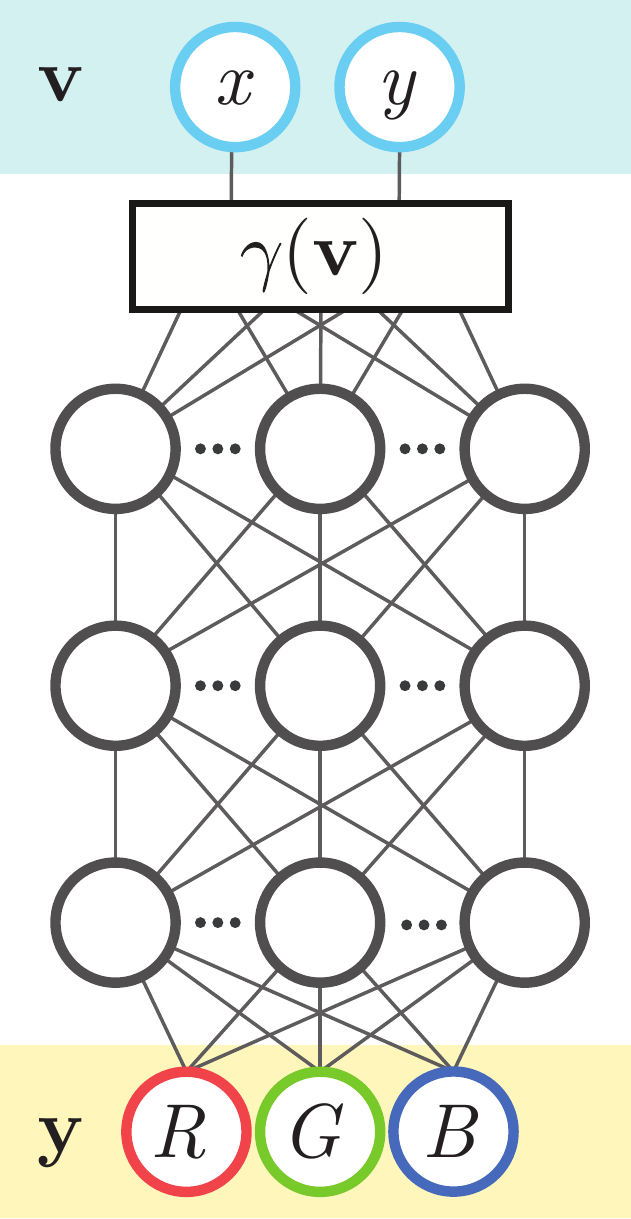} \\
\scriptsize{(a) Coordinate-based MLP}
}
&
\begin{tabular}{@{}c@{\,\,}c@{\,\,}c@{\,\,}c@{\,\,}c@{\,\,}c@{\,\,}c@{}}
\scriptsize\rotatebox{90}{\quad No Fourier features }  &
\scriptsize\rotatebox{90}{\quad\quad\, $\gamma(\vvec) = \vvec$ }  &
\includegraphics[width=0.177\textwidth]{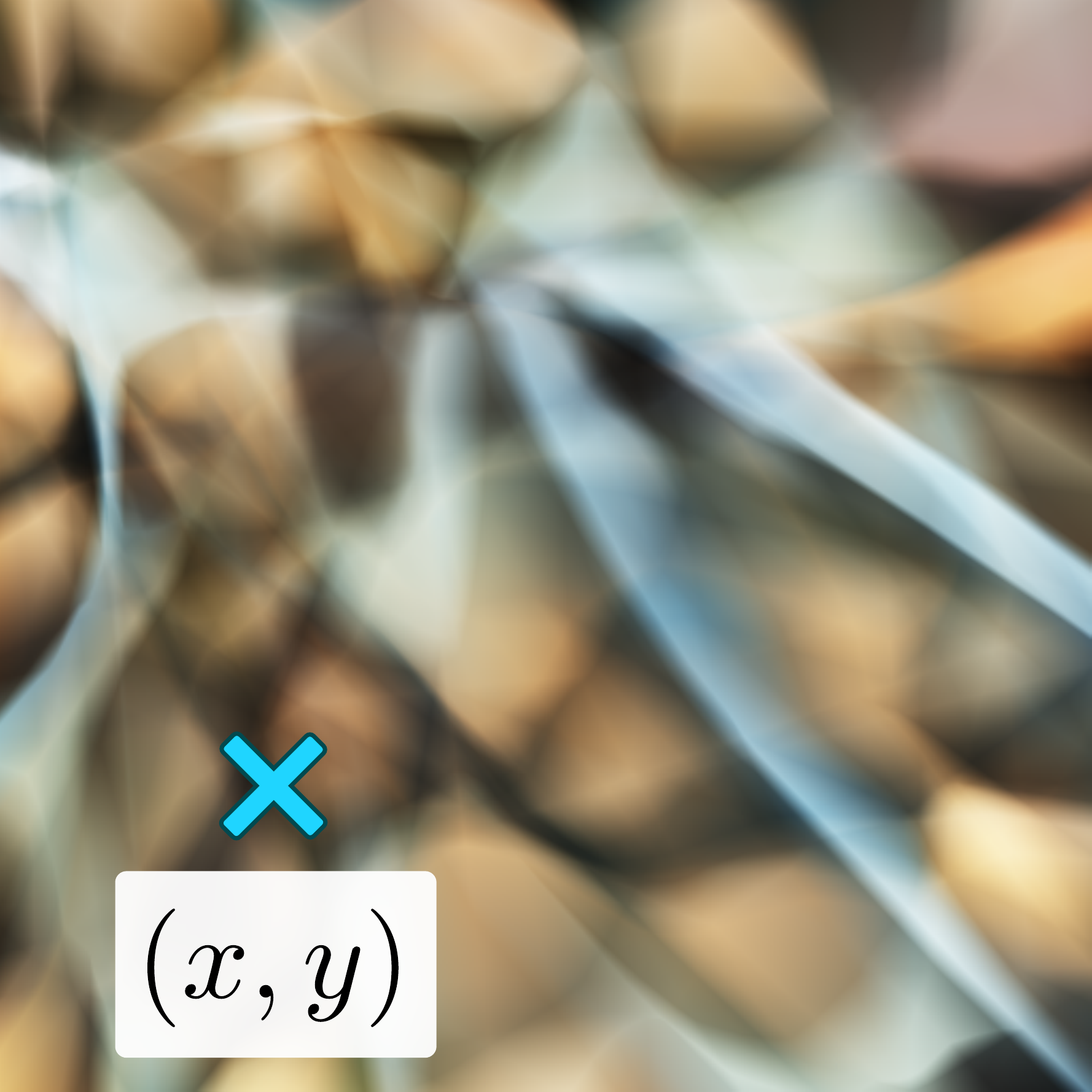} &
\includegraphics[width=0.177\textwidth]{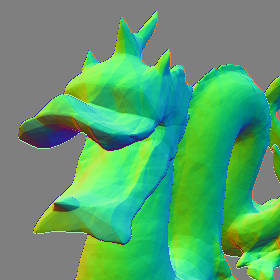} &
\includegraphics[width=0.177\textwidth]{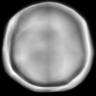} &
\includegraphics[width=0.177\textwidth,trim= 100 100 50 50,clip]{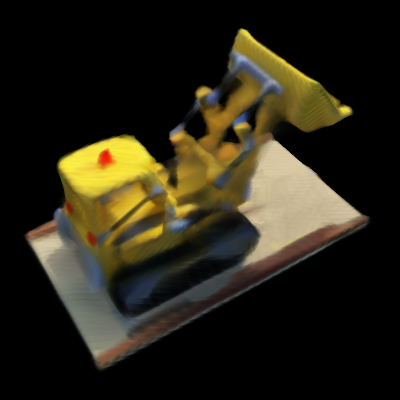} \\
\scriptsize\rotatebox{90}{\,\, With Fourier features }  &
\small\rotatebox{90}{\,\,\, $\gamma(\vvec) = \operatorname{FF}(\vvec)$ }  & \includegraphics[width=0.177\textwidth]{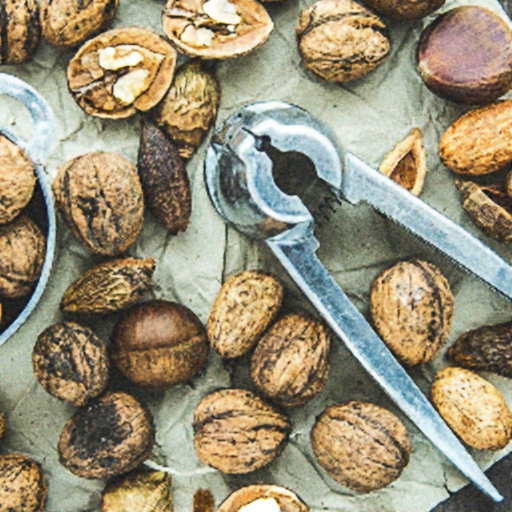} &
\includegraphics[width=0.177\textwidth]{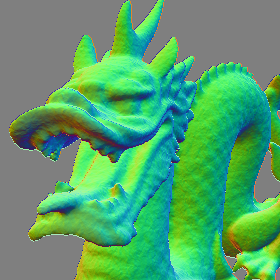} & 
\includegraphics[width=0.177\textwidth]{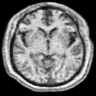} & 
\includegraphics[width=0.177\textwidth,trim= 100 100 50 50,clip]{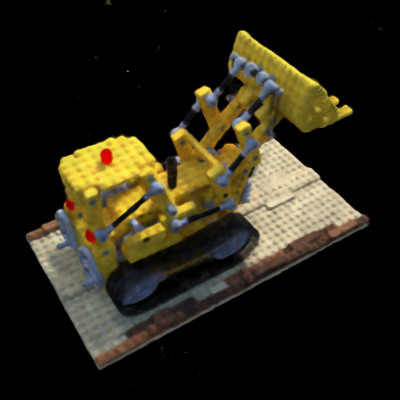} \\
&& \scriptsize{(b) Image regression}
& \scriptsize{(c) 3D shape regression}
& \scriptsize{(d) MRI reconstruction}
& \scriptsize{(e) Inverse rendering} \\
&& \scriptsize{$\!(x,\!y)\!\rightarrow \textrm{RGB}$}
& \scriptsize{$\!(x,\!y,\!z)\!\rightarrow \textrm{occupancy}$}
& \scriptsize{$\!(x,\!y,\!z)\!\rightarrow \textrm{density}$}
& \scriptsize{$\!(x,\!y,\!z)\!\rightarrow\!\textrm{RGB, density}\!$}
\end{tabular}
\end{tabular}
\caption{Fourier features improve the results of coordinate-based MLPs for a variety of high-frequency low-dimensional regression tasks, both with direct (b, c) and indirect (d, e) supervision. We visualize an example MLP (a) for an image regression task (b), where the input to the network is a pixel coordinate and the output is that pixel's color. Passing coordinates directly into the network (top) produces blurry images, whereas preprocessing the input with a Fourier feature mapping (bottom) enables the MLP to represent higher frequency details.
}
\label{fig:teaser}
\end{figure}

A recent line of research in computer vision and graphics replaces traditional discrete representations of objects, scene geometry, and appearance (\eg meshes and voxel grids) with continuous functions parameterized by deep fully-connected networks (also called multilayer perceptrons or MLPs).
These MLPs, which we will call ``coordinate-based'' MLPs, take low-dimensional coordinates as inputs (typically points in $\mathbb{R}^3$) and are trained to output a representation of shape, density, and/or color at each input location (see Figure~\ref{fig:teaser}). 
This strategy is compelling since coordinate-based MLPs are amenable to gradient-based optimization and machine learning, and can be orders of magnitude more compact than grid-sampled representations.
Coordinate-based MLPs have been used to represent images~\cite{nguyen2015deep,stanley2007compositional} (referred to as ``compositional pattern producing networks''), volume density~\cite{mildenhall2020nerf}, occupancy~\cite{occupancynet}, and signed distance~\cite{deepsdf}, and have achieved state-of-the-art results across a variety of tasks such as shape representation~\cite{implicitfields,neuralarticulated,learningshape,localdeep,localimplicit,implicitlayers,deepsdf}, texture synthesis~\cite{henzler2020neuraltexture,texturefields}, shape inference from images~\cite{dist,implicitwithout3d}, and novel view synthesis~\cite{mildenhall2020nerf,diffvolumetric,pifu,srn}.

We leverage recent progress in modeling the behavior of deep networks using kernel regression with a neural tangent kernel (NTK)~\cite{jacot2018neural} to theoretically and experimentally show that standard MLPs are poorly suited for these low-dimensional coordinate-based vision and graphics tasks. In particular, MLPs have difficulty learning high frequency functions, a phenomenon referred to in the literature as ``spectral bias''~\cite{basri2020frequency,rahaman2018spectral}.
NTK theory suggests that this is because standard coordinate-based MLPs correspond to kernels with a rapid frequency falloff, which effectively prevents them from being able to represent the high-frequency content present in natural images and scenes.

A few recent works~\cite{mildenhall2020nerf,Zhong2020Reconstructing} have experimentally found that a heuristic sinusoidal mapping of input coordinates (called a ``positional encoding'') allows MLPs to represent higher frequency content. We observe that this is a special case of Fourier features~\cite{rahimi2007}: mapping input coordinates $\vvec$ to $
\gamma(\vvec) = \Vector{
    a_1 \cos(2 \pi \mathbf b_1^\transpose \vvec),
    a_1 \sin(2 \pi \mathbf b_1^\transpose \vvec), 
    \ldots, 
    a_m \cos(2 \pi \mathbf b_m^\transpose \vvec),
    a_m \sin(2 \pi \mathbf b_m^\transpose \vvec) }$
before passing them into an MLP.
We show that this mapping transforms the NTK into a stationary (shift-invariant) kernel and enables tuning the NTK's spectrum by modifying the frequency vectors $\mathbf b_j$,
thereby controlling the range of frequencies that can be learned by the corresponding MLP.
We show that the simple strategy of setting $a_j=1$ and randomly sampling $\mathbf b_j$
from an isotropic distribution achieves good performance, and that the scale (standard deviation) of this distribution matters much more than its specific shape.
We train MLPs with this Fourier feature input mapping across a range of tasks relevant to the computer vision and graphics communities. As highlighted in Figure~\ref{fig:teaser}, our proposed mapping dramatically improves the performance of coordinate-based MLPs. In summary, we make the following contributions:
\begin{itemize}[leftmargin=*]
\item We leverage NTK theory and simple experiments to show that a Fourier feature mapping can be used to overcome the spectral bias of coordinate-based MLPs towards low frequencies by allowing them to learn much higher frequencies (Section~\ref{sec:fourfeat}).

\item We demonstrate that a random Fourier feature mapping with an appropriately chosen scale can dramatically improve the performance of coordinate-based MLPs across many low-dimensional tasks in computer vision and graphics (Section~\ref{sec:optimization}).

\end{itemize}

\section{Related Work}

Our work is motivated by the widespread use of coordinate-based MLPs to represent a variety of visual signals, including images~\cite{stanley2007compositional} and 3D scenes~\cite{occupancynet,mildenhall2020nerf,deepsdf}. In particular, our analysis is intended to clarify experimental results demonstrating that an input mapping of coordinates (which they called a ``positional encoding'') using sinusoids with logarithmically-spaced axis-aligned frequencies improves the performance of coordinate-based MLPs on the tasks of novel view synthesis from 2D images~\cite{mildenhall2020nerf} and protein structure modeling from cryo-electron microscopy~\cite{Zhong2020Reconstructing}. We analyze this technique to show that it corresponds to a modification of the MLP's NTK, and we show that other non-axis-aligned frequency distributions can outperform this positional encoding. 

Prior works in natural language processing and time series analysis~\cite{kazemi2019time2vec,vaswani2017attention, xu2019selfattention} have used a similar positional encoding to represent time or 1D position. In particular, Xu \etal~\cite{xu2019selfattention} use random Fourier features (RFF)~\cite{rahimi2007} to approximate stationary kernels with a sinusoidal input mapping and propose techniques to tune the mapping parameters. Our work extends this by directly explaining such mappings as a modification of the resulting network's NTK. Additionally, we address the embedding of multidimensional coordinates, which is necessary for vision and graphics tasks.

To analyze the effects of applying a Fourier feature mapping to input coordinates before passing them through an MLP, we rely on recent theoretical work that models neural networks in the limits of infinite width and infinitesimal learning rate as kernel regression using the NTK~\cite{arora2019finegrained,bietti2019inductive,du19,jacot2018neural,lee2019wide}. In particular, we use the analyses from Lee \etal~\cite{lee2019wide} and Arora \etal~\cite{arora2019finegrained}, which show that the outputs of a network throughout gradient descent remain close to those of a linear dynamical system whose convergence rate is governed by the eigenvalues of the NTK matrix~\cite{arora2019finegrained,basri2020frequency,bietti2019inductive,lee2019wide,yang2019fine}. Analysis of the NTK's eigendecomposition shows that its eigenvalue spectrum decays rapidly as a function of frequency, which explains the widely-observed ``spectral bias'' of deep networks towards learning low-frequency functions~\cite{basri2020frequency,basri2019convergence,rahaman2018spectral}.

We leverage this analysis to consider the implications of adding a Fourier feature mapping before the network, and we show that this mapping has a significant effect on the NTK's eigenvalue spectrum and on the corresponding network's convergence properties in practice. 

\section{Background and Notation}
\label{sec:bg}

To lay the foundation for our theoretical analysis, we first review classic kernel regression and its connection to recent results that analyze the training dynamics and generalization behavior of deep fully-connected networks. In later sections, we use these tools to analyze the effects of training coordinate-based MLPs with Fourier feature mappings. 

\shortpara{Kernel regression.}
Kernel regression is a classic nonlinear regression algorithm~\cite{wainwright_2019}. Given a training dataset $(\Data, \mathbf{y}) = \{(\x_i, y_i)\}_{i=1}^\num$, where $\x_i$ are input points and $y_i=f(\x_i)$ are the corresponding scalar output labels, kernel regression constructs an estimate $\hat f$ of the underlying function at any point $\x$ as: 
\begin{equation}
\label{eqn:kernel}
\hat f(\x) = \sum_{i=1}^\num \left( \Kernel^{-1} \mathbf y \right)_i k(\x_i, \x)\,,
\end{equation}
where $\Kernel$ is an $\num\times \num$ kernel (Gram) matrix with entries $\Kernel_{ij} = k(\x_i, \x_j)$ and $k$ is a symmetric positive semidefinite (PSD) kernel function which represents the ``similarity'' between two input vectors. Intuitively, the kernel regression estimate at any point $\x$ can be thought of as a weighted sum of training labels $y_i$ using the similarity between the corresponding $\x_i$ and $\x$.

\shortpara{Approximating deep networks with kernel regression.} Let $f$ be a fully-connected deep network with weights $\theta$ initialized from a Gaussian distribution $\N$.
Theory proposed by Jacot \etal~\cite{jacot2018neural} and extended by others~\cite{arora2019finegrained,basri2020frequency,lee2019wide} shows that when the width of the layers in $f$ tends to infinity and the learning rate for SGD tends to zero, the function $f(\mathbf x; \theta)$ converges over the course of training to the kernel regression solution using the \emph{neural tangent kernel} (NTK), defined as: 
\begin{equation}
\label{eqn:ntk}
    k_{\mathrm{NTK}}(\x_i, \x_j) = \mathbb E_{\theta \sim \N} \left\langle \frac{\partial f(\x_i; \theta)}{\partial \theta}, \frac{\partial f(\x_j; \theta)}{\partial \theta} \right\rangle\,.
\end{equation}
When the inputs are restricted to a hypersphere, the NTK for an MLP can be written as a dot product kernel (a kernel in the form $\hNTK(\mathbf x_i^\transpose \mathbf x_j)$ for a scalar function $\hNTK : \mathbb R \to \mathbb R$).

Prior work~\cite{arora2019finegrained,basri2020frequency,jacot2018neural,lee2019wide} shows that an NTK linear system model can be used to approximate the dynamics of a deep network during training. We consider a network trained with an \ltwo loss and a learning rate $\eta$, where the network's weights are initialized such that the output of the network at initialization is close to zero. Under asymptotic conditions stated in Lee \etal~\cite{lee2019wide}, the network's output for any data $\arbitrary{\Data}$ after $t$ training iterations can be approximated as:
\begin{equation}
\hat{\mathbf{y}}^{(t)} \approx \arbitrary{\Kernel} \Kernel^{-1}\left( \Eye - e^{-\eta \Kernel t} \right) \mathbf{y}\,,
\label{eqn:testvals}
\end{equation}
where $\hat{\mathbf{y}}^{(t)}=f(\arbitrary{\Data}; \theta)$ are the network's predictions on input points $\arbitrary{\Data}$ at training iteration $t$, $\Kernel$ is the NTK matrix between all pairs of training points in $\Data$, and $\arbitrary{\Kernel}$ is the NTK matrix between all points in $\arbitrary{\Data}$ and all points in the training dataset $\Data$.

\shortpara{Spectral bias when training neural networks.}
Let us consider the training error $\mathbf{\hat{y}}^{(t)}_\textrm{train}-\mathbf{y}$, where $\mathbf{\hat{y}}^{(t)}_\textrm{train}$ are the network's predictions on the training dataset at iteration $t$. Since the NTK matrix $\Kernel$ must be PSD, we can take its eigendecomposition $\Kernel = \Eigvec \Eigval \Eigvec^\transpose$, where $\Eigvec$ is orthogonal and $\Eigval$ is a diagonal matrix whose entries are the eigenvalues $\lambda_i \geq 0$ of $\Kernel$.
Then, since $e^{- \eta \Kernel t} = \Eigvec e^{- \eta \Eigval t} \Eigvec^\transpose$:
\begin{equation}
    \Eigvec^\transpose (\mathbf{\hat{y}}^{(t)}_\textrm{train} - \mathbf y) 
    \approx \Eigvec^\transpose \left(\left( \Eye - e^{-\eta \Kernel t} \right) \mathbf{y} - \mathbf{y} \right)  
    = -e^{-\eta \Eigval t} \Eigvec^\transpose \mathbf{y}\,.
\label{eqn:eig}
\end{equation}
This means that if we consider training convergence in the eigenbasis of the NTK, the $i^{\textrm{th}}$ component of the absolute error $ | \Eigvec^\transpose (\mathbf{\hat{y}}^{(t)}_\textrm{train} - \mathbf y)  |_i$ will decay approximately exponentially at the rate $\eta \lambda_i$. In other words, components of the target function that correspond to kernel eigenvectors with larger eigenvalues will be learned faster. For a conventional MLP, the eigenvalues of the NTK decay rapidly \cite{basri2019convergence,bietti2019inductive, heckel2020compressive}. This results in extremely slow convergence to the high frequency components of the target function, to the point where standard MLPs are effectively unable to learn these components, as visualized in Figure~\ref{fig:teaser}. Next, we describe a technique to address this slow convergence by using a Fourier feature mapping of input coordinates before passing them to the MLP.

\section{Fourier Features for a Tunable Stationary Neural Tangent Kernel}
\label{sec:fourfeat}

Machine learning analysis typically addresses the case in which inputs are high dimensional points (\eg the pixels of an image reshaped into a vector) and training examples are sparsely distributed.
In contrast, in this work we consider \emph{low-dimensional regression} tasks, wherein inputs are assumed to be dense coordinates in a subset of $\mathbb R^d$ for small values of $d$ (\eg pixel coordinates).
%(roughly, $d \leq 5$).
This setting has two significant implications when viewing deep networks through the lens of kernel regression:
\begin{enumerate}[leftmargin=*]
    \item We would like the composed NTK to be shift-invariant over the input domain, since the training points are distributed with uniform density. In problems where the inputs are normalized to the surface of a hypersphere (common in machine learning), a dot product kernel (such as the regular NTK) corresponds to spherical convolution. However, inputs in our setting are dense in Euclidean space. A Fourier feature mapping of input coordinates makes the composed NTK stationary (shift-invariant), acting as a convolution kernel over the input domain (see Appendix~\ref{sec:stationarity} for additional discussion on stationary kernels).
    \item We would like to control the bandwidth of the NTK to improve training speed and generalization. As we see from Eqn.~\ref{eqn:eig}, a ``wider'' kernel with a slower spectral falloff achieves faster training convergence for high frequency components. However, we know from signal processing that reconstructing a signal using a kernel whose spectrum is \emph{too} wide causes high frequency aliasing artifacts. We show in Section~\ref{sec:optimization} that a Fourier feature input mapping can be tuned to lie between these ``underfitting' and ``overfitting'' extremes, enabling both fast convergence and low test error. 
\end{enumerate}

\shortpara{Fourier features and the composed neural tangent kernel.} 
Fourier feature mappings have been used in many applications since their introduction in the seminal work of Rahimi and Recht~\cite{rahimi2007}, which used random Fourier features to approximate an arbitrary stationary kernel function by applying Bochner's theorem. Extending this technique, we use a Fourier feature mapping $\gamma$ to featurize input coordinates before passing them through a coordinate-based MLP, and investigate the theoretical and practical effect this has on convergence speed and generalization. The function $\gamma$ maps input points $\vvec \in [0,1)^d$ to the surface of a higher dimensional hypersphere with a set of sinusoids:
\begin{equation}
\label{eqn:gamma}
    \gamma(\vvec) = \Vector{
    a_1 \cos(2 \pi \mathbf b_1^\transpose \vvec),
    a_1 \sin(2 \pi \mathbf b_1^\transpose \vvec), 
    \ldots, 
    a_m \cos(2 \pi \mathbf b_m^\transpose \vvec),
    a_m \sin(2 \pi \mathbf b_m^\transpose \vvec) } \,.
\end{equation}
Because $\cos(\alpha-\beta) = \cos \alpha \cos \beta + \sin \alpha \sin \beta$, the kernel function induced by this mapping is:
\begin{gather}
    k_\gamma(\vvec_1, \vvec_2) = \gamma(\vvec_1)^\transpose \gamma(\vvec_2) 
    = \sum_{j=1}^m a_j^2 \cos\left(2 \pi \mathbf b_j^\transpose \left(\vvec_1 - \vvec_2 \right) \right)
    = h_\gamma(\vvec_1 - \vvec_2)\,, \\
    \textrm{where } h_\gamma(\vvec_\Delta) \triangleq \sum_{j=1}^m a_j^2 \cos(2 \pi \mathbf b_j^\transpose \vvec_\Delta) \, .
\end{gather}
Note that this kernel is stationary (a function of only the difference between points). We can think of the mapping as a Fourier approximation of a kernel function: $\mathbf b_j$ are the Fourier basis frequencies used to approximate the kernel, and $a_j^2$ are the corresponding Fourier series coefficients.

After computing the Fourier features for our input points, we pass them through an MLP to get $f(\gamma(\vvec) ; \theta)$. As discussed previously, the result of training a network can be approximated by kernel regression using the kernel $\hNTK(\x_i^\transpose \x_j)$. In our case, $\x_i = \gamma(\vvec_i)$ so the composed kernel becomes:
\begin{equation}
    \hNTK(\x_i^\transpose \x_j) = \hNTK\left(\gamma\left(\vvec_i\right)^\transpose \gamma\left(\vvec_j\right)\right)  = \hNTK \left(h_\gamma \left(\vvec_i - \vvec_j \right) \right).
\end{equation}
Thus, training a network on these embedded input points corresponds to kernel regression with the \emph{stationary} composed NTK function $\hNTK \circ h_\gamma$\,. The MLP function approximates a convolution of the composed NTK with a weighted Dirac delta at each input training point $\vvec_i$:
\begin{align}
    \hat f = \left( \hNTK \circ h_\gamma \right) * \sum_{i=1}^\num w_i \delta_{\vvec_i}
\end{align}
where $\mathbf w = \Kernel^{-1} \mathbf y$ (from Eqn.~\ref{eqn:kernel}).
This allows us to draw analogies to signal processing, where the composed NTK acts similarly to a reconstruction filter. In the next section, we show that the frequency decay of the composed NTK determines the behavior of the reconstructed signal.

\section{Manipulating the Fourier Feature Mapping}
\label{sec:optimization}

Preprocessing the inputs to a coordinate-based MLP with a Fourier feature mapping creates a composed NTK that is not only stationary but also \emph{tunable}. By manipulating the settings of the $a_j$ and $\mathbf b_j$ parameters in Eqn.~\ref{eqn:gamma}, it is possible to dramatically change both the rate of convergence and the generalization behavior of the resulting network.
In this section, we investigate the effects of the Fourier feature mapping in the setting of 1D function regression.

We train MLPs to learn signals $f$ defined on the interval $[0,1)$. We sample $c \num$ linearly spaced points on the interval, 
using every $c^\textrm{th}$ point as the training set and the remaining points as the test set.
Since our composed kernel function is stationary, evaluating it at linearly spaced points on a periodic domain makes the resulting kernel matrix circulant: it represents a convolution and is diagonalizable by the Fourier transform. Thus, we can compute the eigenvalues of the composed NTK matrix by simply taking the Fourier transform of a single row. All experiments are implemented in JAX~\cite{jax2018github} and the NTK functions are calculated automatically using the Neural Tangents library~\cite{neuraltangents2020}.

\begin{figure}[t]
    \centering
    \includegraphics[width=\textwidth]{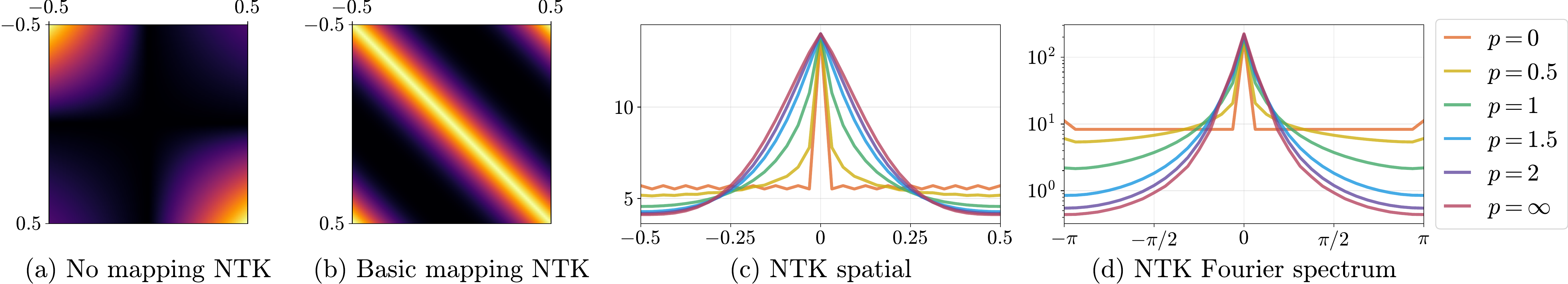}
    \caption{
    Adding a Fourier feature mapping can improve the poor conditioning of a coordinate-based MLP's neural tangent kernel (NTK).
    (a) We visualize the NTK function $k_{\mathrm{NTK}}(x_i, x_j)$ (Eqn.~\ref{eqn:ntk}) for a 4-layer ReLU MLP with one scalar input.
    This kernel is not shift-invariant and does not have a strong diagonal, making it poorly suited for kernel regression in low-dimensional problems.
    (b) A basic input mapping $\gamma(v) = \Vector{ \cos 2\pi v, \sin 2\pi v }$ 
    makes the composed NTK $k_{\mathrm{NTK}}(\gamma(v_i), \gamma(v_j))$ shift-invariant (stationary). 
    (c) A Fourier feature input mapping (Eqn.~\ref{eqn:gamma}) can be used to tune the composed kernel's width, where we set $a_j = 1/j^p$ and $b_j = j$ for $j = 1,\ldots, \num/2$. (d) Higher frequency mappings (lower $p$) result in composed kernels with wider spectra, which enables faster convergence for high-frequency components (see Figure~\ref{fig:1d_dense}).
    }
    \label{fig:ntks}
\end{figure}

\shortpara{Visualizing the composed NTK.}
We first visualize how modifying the Fourier feature mapping changes the composed NTK. We set $b_j = j$ (full Fourier basis in 1D) and $a_j = 1/j^p$ for $j = 1,\ldots, \num/2$. 
We use $p=\infty$ to denote the mapping $\gamma(v) = \Vector{ \cos 2\pi v, \sin 2\pi v }$ that simply wraps $[0,1)$ around the unit circle (this is referred to as the ``basic'' mapping in later experiments).
Figure~\ref{fig:ntks} demonstrates the effect of varying $p$ on the composed NTK. By construction, lower $p$ values result in a slower falloff in the frequency domain and a correspondingly narrower kernel in the spatial domain.

\shortpara{Effects of Fourier features on network convergence.}
We generate ground truth 1D functions by sampling $c\num$ values from a family with parameter $\alpha$ as follows: we sample a standard i.i.d. Gaussian vector of length $c\num$, scale its $i^\textrm{th}$ entry by $1/i^{\alpha}$, then return the real component of its inverse Fourier transform. We will refer to this as a ``$1/f^\alpha$ noise'' signal.

\begin{figure}[t]
    \centering
    \includegraphics[width=\textwidth]{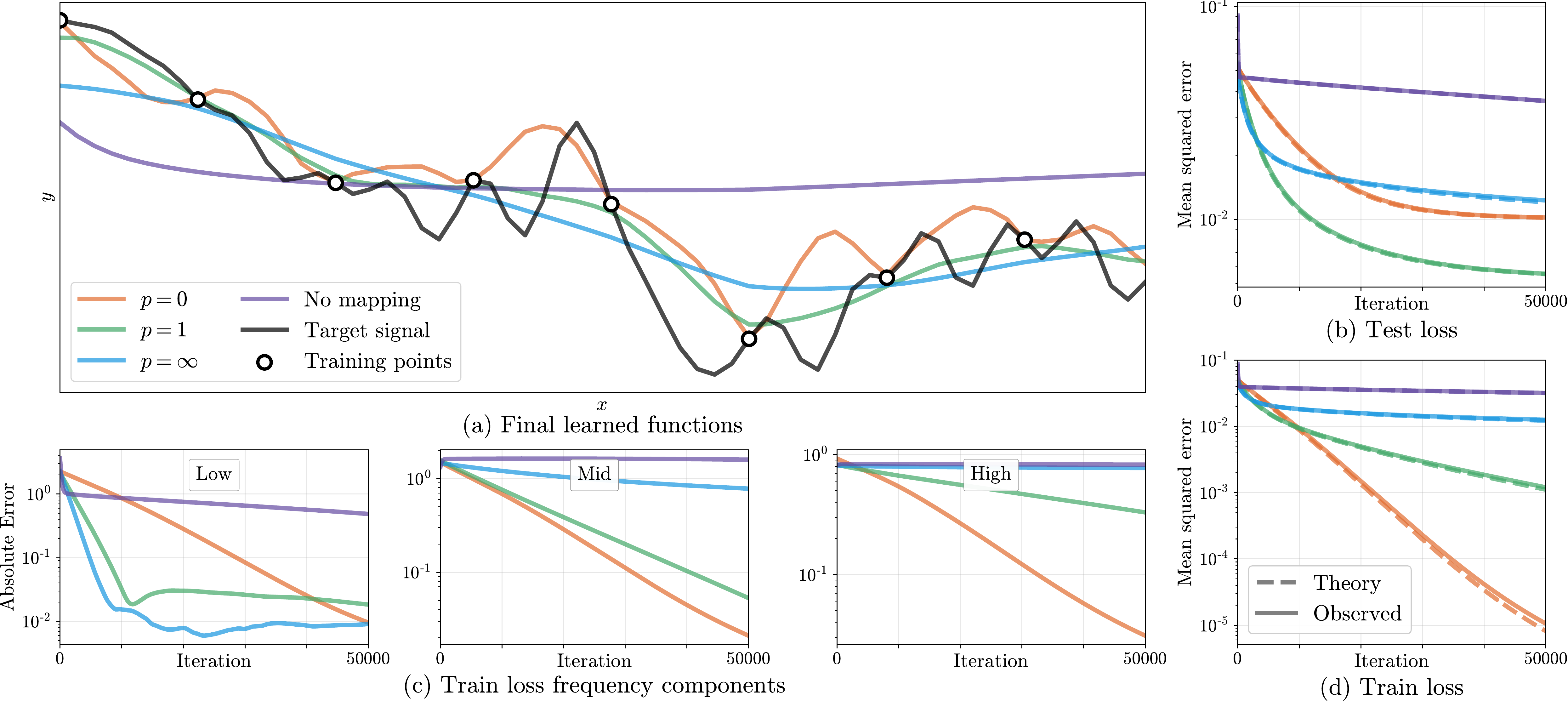}
    \caption{Combining a network with a Fourier feature mapping has dramatic effects on convergence and generalization. 
    Here we train a network on 32 sampled points from a 1D function (a) using mappings shown in Fig.~\ref{fig:ntks}. A mapping with a smaller $p$ value yields a composed NTK with more power in higher frequencies, enabling the corresponding network to learn a higher frequency function.
    The theoretical and experimental training loss improves monotonically with higher frequency kernels (d), but the test-set loss is lowest at $p=1$ and falls as the network starts to overfit (b). As predicted by Eqn.~\ref{eqn:eig}, we see roughly log-linear convergence of the training loss frequency components (c). Higher frequency kernels result in faster convergence for high frequency loss components, thereby overcoming the ``spectral bias'' observed when training networks with no input mapping.}
    \label{fig:1d_dense}
\end{figure}

In Figure~\ref{fig:1d_dense}, we train MLPs (4 layers, 1024 channels, ReLU activations) to fit a bandlimited $1/f^1$ noise signal ($c=8,n=32$) using Fourier feature mappings with different $p$ values. Figures~\ref{fig:1d_dense}b and \ref{fig:1d_dense}d show that the NTK linear dynamics model accurately predict the effects of modifying the Fourier feature mapping parameters. Separating different frequency components of the training error in Figure~\ref{fig:1d_dense}c reveals that networks with narrower NTK spectra converge faster for low frequency components but essentially never converge for high frequency components, whereas networks with wider NTK spectra successfully converge across all components. 
The Fourier feature mapping $p=1$ has adequate power across frequencies present in the target signal (so the network converges rapidly during training) but limited power in higher frequencies (preventing overfitting or aliasing).

\shortpara{Tuning Fourier features in practice.}
Eqn.~\ref{eqn:testvals} allows us to estimate a trained network's theoretical loss on a validation set using the composed kernel. For small 1D problems, we can minimize this loss with gradient-based optimization to choose mapping parameters $a_j$ (given a dense sampling of $b_j$). 
In this carefully controlled setting (1D signals, small training dataset, gradient descent with small learning rate, very wide networks), we find that this optimized mapping also achieves the best performance when training networks.
Please refer to Appendix~\ref{sec:supp_opt_ntk} for details and experiments. 

In real-world problems, especially in multiple dimensions, it is not feasible to use a feature mapping that densely samples Fourier basis functions; the number of Fourier basis functions scales with the number of training data points, which grows exponentially with dimension. Instead, we sample a set of random Fourier features~\cite{rahimi2007} from a parametric distribution. We find that the exact sampling distribution family is much less important than the distribution's scale (standard deviation).

\begin{figure}[t]
    \centering
    \includegraphics[width=\textwidth]{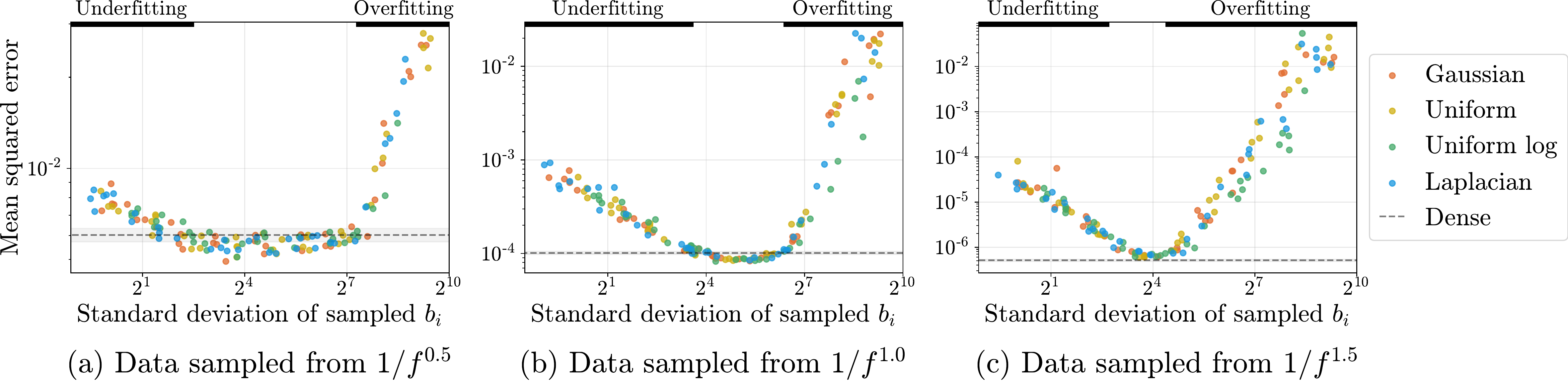}
    \caption{
    We find that a sparse random sampling of Fourier features can perform as well as a dense set of features and that the width of the distribution matters more than the shape.
    Here, we generate random 1D signals from $1/f^\alpha$ noise and report the test-set accuracy of different trained models that use a sparse set (16 out of 1024) of random Fourier features sampled from different distributions. Each subplot represents a different family of 1D signals.
    Each dot represents a trained network, where the color indicates which Fourier feature sampling distribution is used. We plot the test error of each model versus the empirical standard deviation of its sampled frequencies.
    The best models using sparsely sampled features are able to match the performance of a model trained with dense Fourier features (dashed lines with error bars). 
    All sampling distributions trace out the same curve, exhibiting underfitting (slow convergence) when the standard deviation of sampled frequencies is too low and overfitting when it is too high.
    This implies that the precise shape of the distribution used to sample frequencies does not have a significant impact on performance.
    }
    \label{fig:1d_sparse}
\end{figure}

Figure~\ref{fig:1d_sparse} demonstrates this point using hyperparameter sweeps for a variety of sampling distributions. In each subfigure, we draw 1D target signals ($c=2,n=1024$) from a fixed $1/f^\alpha$ distribution and train networks to learn them. We use random Fourier feature mappings (of length 16) sampled from different distribution families (Gaussian, uniform, uniform in log space, and Laplacian) and sweep over each distribution's scale. 
Perhaps surprisingly, the standard deviation of the sampled frequencies alone is enough to predict test set performance, regardless of the underlying distribution's shape. We show that this holds for higher-dimensional tasks in Appendix~\ref{sec:supp_2d_scatter}. We also observe that passing this sparse sampling of Fourier features through an MLP matches the performance of using a dense set of Fourier features with the same MLP, suggesting a strategy for scaling to higher dimensions. We proceed with a Gaussian distribution for our higher-dimensional experiments in Section~\ref{sec:experiments} and treat the scale as a hyperparameter to tune on a validation dataset.

\section{Experiments}
\label{sec:experiments}

We validate the benefits of using Fourier feature mappings for coordinate-based MLPs with experiments on a variety of regression tasks relevant to the computer vision and graphics communities.

\subsection{Compared mappings}

In Table~\ref{table:results}, we compare the performance of coordinate-based MLPs with no input mapping and with the following Fourier feature mappings ($\cos,\sin$ are applied elementwise):

\shortpara{Basic:} 
$\gamma(\vvec)=\Vector{\cos(2 \pi \vvec{v}),\sin(2 \pi \vvec)}$. 
Simply wraps input coordinates around the circle.

\shortpara{Positional encoding:} 
$\gamma(\vvec)=\Vector{\ldots, \cos(2 \pi \sigma^{j/m} \vvec),\sin(2 \pi \sigma^{j/m} \vvec), \ldots}$ for $j = 0, \ldots, m-1$. 
Uses log-linear spaced frequencies for each dimension, where the scale $\sigma$ is chosen for each task and dataset by a hyperparameter sweep. This is a generalization of the ``positional encoding'' used by prior work~\cite{mildenhall2020nerf,vaswani2017attention,Zhong2020Reconstructing}. Note that this mapping is deterministic and only contains on-axis frequencies, making it naturally biased towards data that has more frequency content along the axes.

\shortpara{Gaussian:}
$\gamma(\vvec)=\Vector{\cos(2 \pi \mathbf B \vvec), \sin(2 \pi \mathbf B \vvec)}$, 
where each entry in $\mathbf B \in \mathbb R^{m \times d}$ is sampled from $\mathcal N(0,\sigma^2)$, and $\sigma$ is chosen for each task and dataset with a hyperparameter sweep. 
In the absence of any strong prior on the frequency spectrum of the signal, we use an isotropic Gaussian distribution.

Our experiments show that all of the Fourier feature mappings improve the performance of coordinate-based MLPs over using no mapping and that the Gaussian RFF mapping performs best.

\subsection{Tasks}

We conduct experiments with direct regression, where supervision labels are in the same space as the network outputs, as well as indirect regression, where the network outputs are passed through a forward model to produce observations in the same space as the supervision labels (Appendix~\ref{sec:linear map} contains a theoretical analysis of indirect regression through a linear forward model). For each task and dataset, we tune Fourier feature scales on a held-out set of signals. 
For each target signal, we train an MLP on a training subset of the signal and compute error over the remaining test subset.
All tasks (except 3D shape regression) use \ltwo loss and a ReLU MLP with 4 layers and 256 channels. The 3D shape regression task uses cross-entropy loss and a ReLU MLP with 8 layers and 256 channels. We apply a sigmoid activation to the output for each task (except the view synthesis density prediction). We use 256 frequencies for the feature mapping in all experiments (see Appendix~\ref{sec:supp_sparsity} for experiments that investigate the effects of network depth and feature mapping sparsity). Appendix~\ref{sec:supp_tasks} provides additional details on each task and our implementations, and Appendix~\ref{sec:supp_images} shows more result figures.

\shortpara{2D image regression.}
In this task, we train an MLP to regress from a 2D input pixel coordinate to the corresponding RGB value of an image. 
For each test image, we train an MLP on a regularly-spaced grid containing $\nicefrac{1}{4}$ of the pixels and report test error on the remaining pixels.
We compare input mappings over a dataset of natural images and a dataset of text images.

\shortpara{3D shape regression.}
Occupancy Networks~\cite{occupancynet} implicitly represent a 3D shape as the ``decision boundary'' of an MLP, which is trained to output 0 for points outside the shape and 1 for points inside the shape. Each batch of training data is generated by sampling points uniformly at random from the bounding box of the shape and calculating their labels using the ground truth mesh. Test error is calculated using intersection-over-union versus ground truth on a set of points randomly sampled near the mesh surface to better highlight the different mappings' abilities to resolve fine details.

%%%%%%%

\shortpara{2D computed tomography (CT).}
In CT, we observe integral projections of a density field instead of direct measurements. In our 2D CT experiments, we train an MLP that takes in a 2D pixel coordinate and predicts the corresponding volume density at that location. The network is indirectly supervised by the loss between a sparse set of ground-truth integral projections and integral projections computed from the network's output. We conduct experiments using two datasets: procedurally-generated Shepp-Logan phantoms~\cite{shepp} and 2D brain images from the ATLAS dataset~\cite{atlas}.

\shortpara{3D magnetic resonance imaging (MRI).}
In MRI, we observe Fourier transform coefficients of atomic response to radio waves under a magnetic field. In our 3D MRI experiments, we train an MLP that takes in a 3D voxel coordinate and predicts the corresponding response at that location. The network is indirectly supervised by the loss between a sparse set of ground-truth Fourier transform coefficients and Fourier transform coefficients computed from discretely querying the MLP on a voxel grid. We conduct experiments using the ATLAS dataset~\cite{atlas}.

\shortpara{3D inverse rendering for view synthesis.}
In view synthesis, we observe 2D photographs of a 3D scene, reconstruct a representation of that scene, then render images from new viewpoints. To perform this task, we train a coordinate-based MLP that takes in a 3D location and outputs a color and volume density. This MLP is indirectly supervised by the loss between the set of 2D image observations and the same viewpoints re-rendered from the predicted scene representation. We use a simplified version of the method described in NeRF~\cite{mildenhall2020nerf}, where we remove hierarchical sampling and view dependence and replace the original positional encoding with our compared input mappings.

\begin{table}[t]
\centering
\resizebox{\textwidth}{!}{%
\begin{tabular}{l||c@{\quad}c|c|c@{\quad}c|c|c}
             & \multicolumn{3}{c|}{Direct supervision} & \multicolumn{4}{c}{Indirect supervision}  \\  
            & \multicolumn{2}{c|}{2D image} & 3D shape~\cite{occupancynet} & \multicolumn{2}{c|}{2D CT} & \multicolumn{1}{c|}{3D MRI} & 3D NeRF~\cite{mildenhall2020nerf} \\ 
            & Natural & Text & & Shepp & ATLAS & ATLAS &  \\ \hline
           No mapping      & 19.32& 18.40& 0.864& 16.75& 15.44& 26.14& 22.41 \\
           Basic           & 21.71& 20.48& 0.892& 23.31& 16.95& 28.58& 23.16\\ 
           Positional enc. & 24.95& 27.57& 0.960& 26.89& 19.55& 32.23& 25.28\\
           Gaussian        & \textbf{25.57}& \textbf{30.47}& \textbf{0.973}& \textbf{28.33}& \textbf{19.88}& \textbf{34.51}& \textbf{25.48}
\end{tabular}
}
\vspace{2mm}
\caption{We compare four different input mappings on a variety of low-dimensional regression tasks. All results are reported in PSNR except \textit{3D shape}, which uses IoU (higher is better for all). 
\textit{No mapping} represents using a standard MLP with no feature mapping. 
\textit{Basic}, \textit{Positional encoding}, and \textit{Gaussian} are different variants of Fourier feature maps.
For the \textit{Direct supervision} tasks, the network is supervised using ground truth labels for each input coordinate. For the \textit{Indirect supervision} tasks, the network outputs are passed through a forward model before the loss is applied (integral projection for CT, the Fourier transform for MRI, and nonlinear volume rendering for NeRF).
Fourier feature mappings improve results across all tasks, with random Gaussian features performing best.
}
\vspace{-2mm}
\label{table:results}
\end{table}

\section{Conclusion}
We leverage NTK theory to show that a Fourier feature mapping can make coordinate-based MLPs better suited for modeling functions in low dimensions, thereby overcoming the spectral bias inherent in coordinate-based MLPs. We experimentally show that tuning the Fourier feature parameters offers control over the frequency falloff of the combined NTK and significantly improves performance across a range of graphics and imaging tasks. 
These findings shed light on the burgeoning technique of using coordinate-based MLPs to represent 3D shapes in computer vision and graphics pipelines, and provide a simple strategy for practitioners to improve results in these domains. 

\section*{Acknowledgements}

We thank Ben Recht for advice, and Cecilia Zhang and Tim Brooks for their comments on the text.
BM is funded by a Hertz Foundation Fellowship and acknowledges support from the Google BAIR Commons program. 
MT, PS and SFK are funded by NSF Graduate Fellowships.
RR was supported in part by ONR grants N000141712687 and
N000142012529 and the Ronald L. Graham Chair.
RN was supported in part by an FHL Vive Center Seed Grant.
Google University Relations provided a generous donation of compute credits.

\bibliographystyle{plain}
\bibliography{references}

\newcommand{\picwidth}{0.19\textwidth}

\begin{appendices}

\renewcommand{\contentsname}{Appendix Table of Contents}
% \tableofcontents

\section{Further experiments}

\subsection{Optimizing validation error through the NTK linear dynamics}
\label{sec:supp_opt_ntk}

Using Eqn.~\ref{eqn:testvals} in the main paper, we can predict what error a trained network will achieve on a set of testing points. Since this equation depends on the composed NTK, we can directly relate predicted test set loss to the Fourier feature mapping parameters $a$ and $b$ for a validation set of signals $\mathbf y_{val}$:
\begin{equation}
\mathcal{L}_\opt=\norm{\mathbf{u}^{(t)}-\mathbf{y}_\val}_2^2\approx\norm{\Kernel_\val \Kernel^{-1} \left(\Eye- e^{-\eta \Kernel t} \right) \mathbf{y} - \mathbf{y}_\val^{\phantom{^{(t)}}}}_2^2,
\label{eqn:lopt}
\end{equation}
where $\Kernel_\val$ is the composed NTK evaluated between points in a validation dataset $\Data_\val$ and training dataset $\Data$, and $\eta$ and $t$ are the learning rate and number of iterations that will be used when training the actual network.

\begin{figure}[h!]
    \centering
    \includegraphics[width=\textwidth]{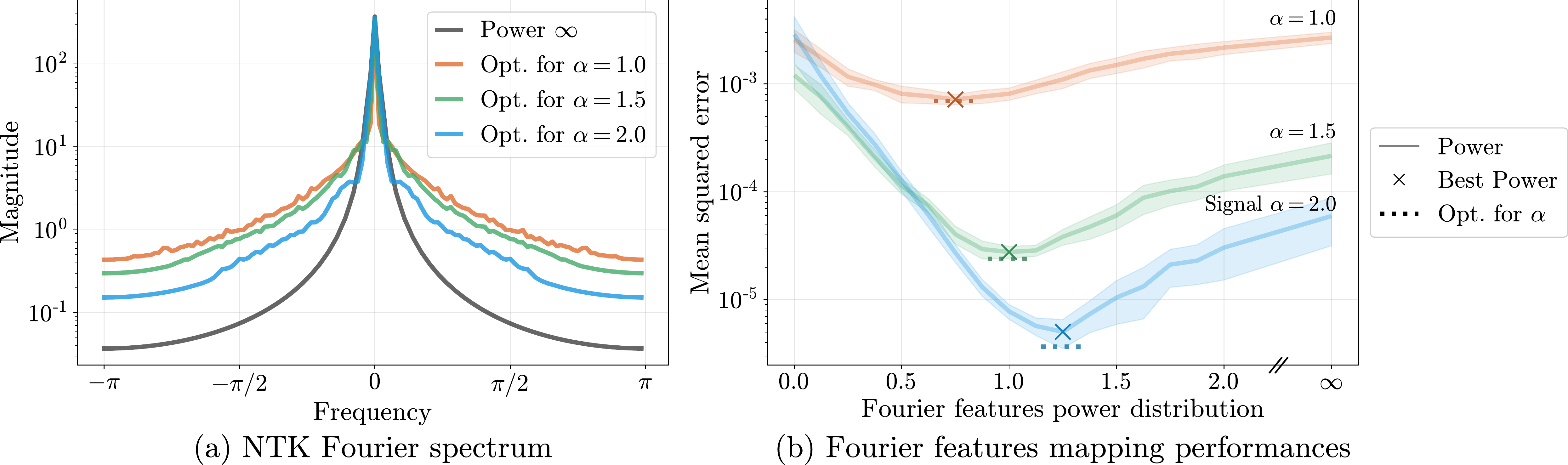}
    \caption{The Fourier feature mappings can be optimized for better performance on a class of target signals by using the linearized network approximation. Here we consider target signals sampled from three different power law distributions. In (a) we show the spectrum for composed kernels corresponding to different optimized feature mappings, where the feature mappings are initialized to match the ``Power $\infty$'' distribution. In (b) we take an alternative approach where we sweep over "power law" settings for our Fourier features. We find that tuning this simple parameterization is able to perform on par with the optimized feature maps.
    }
    \label{fig:1d_opt}
\end{figure}

In Figure~\ref{fig:1d_opt}, we show the results of minimizing Eqn.~\ref{eqn:lopt} by gradient descent on $a_j$ values (with fixed corresponding ``densely sampled'' $b_j=j$) for validation sets sampled from three different $1/f^\alpha$ noise families. Note that gradient descent on this theoretical loss approximation produces $a_j$ values which are able to perform as well as the best ``power law'' $a_j$ values for each respective signal class (compared dashed lines versus $\times$ markers in Figure~\ref{fig:1d_opt}b). As mentioned in the main text, we find that this optimization strategy is only viable for small 1D regression problems. In our multidimensional tasks, using densely sampled $\mathbf b_j$ values is not tractable due to memory constraints. In addition, the theoretical approximation only holds when training the network using SGD, and in practice we train using the Adam optimizer~\cite{adam}.

\subsection{Feature sparsity and network depth}
\label{sec:supp_sparsity}

\begin{figure}[h!]
  \begin{minipage}[c]{0.48\textwidth}
    \includegraphics[width=\textwidth]{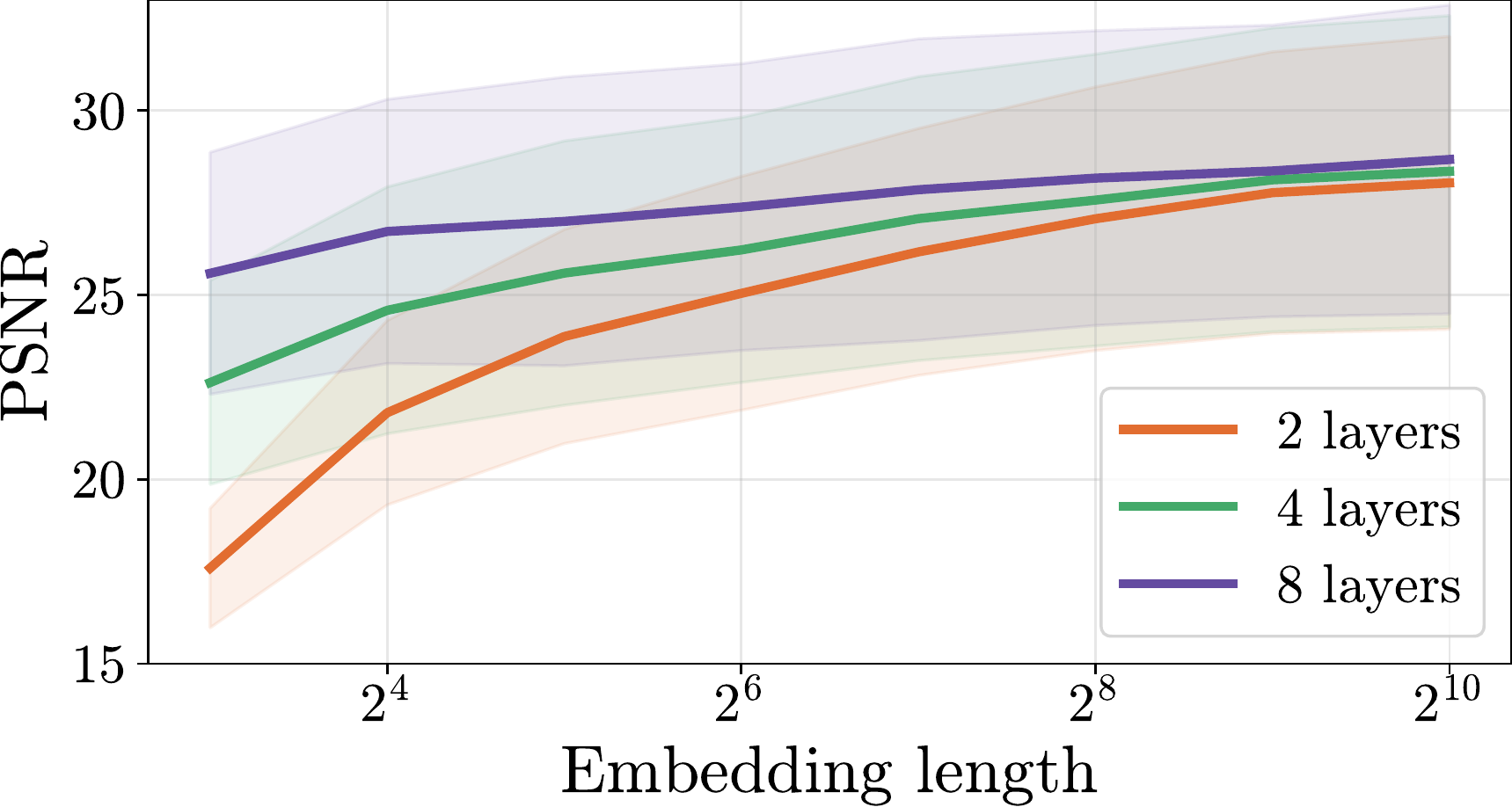}
  \end{minipage}\hfill
  \begin{minipage}[c]{0.48\textwidth}
    \caption{In a 2D image regression task (explained in Section~\ref{sec:2d_image}) we find that shallower networks require more Fourier features than deeper networks. This is explained by the frequency spreading effect shown in Figure~\ref{fig:frequency_spreading}. In this experiment we use the \emph{Natural} image dataset and a Gaussian mapping. All of the network layers have 256 channels, and the networks are trained using an Adam~\cite{adam} optimizer with a learning rate of $10^{-3}$.}
    \label{fig:2d_sparse}
  \end{minipage}
\end{figure}

In our experiments, we observe that deeper networks need fewer Fourier features than shallow networks. As the depth of the MLP increases, we observe that a sparser set of frequencies can achieve similar performance; Figure~\ref{fig:2d_sparse} illustrates this effect in the context of 2D image regression.

\begin{figure}[h!]
    \centering
    \includegraphics[width=\textwidth]{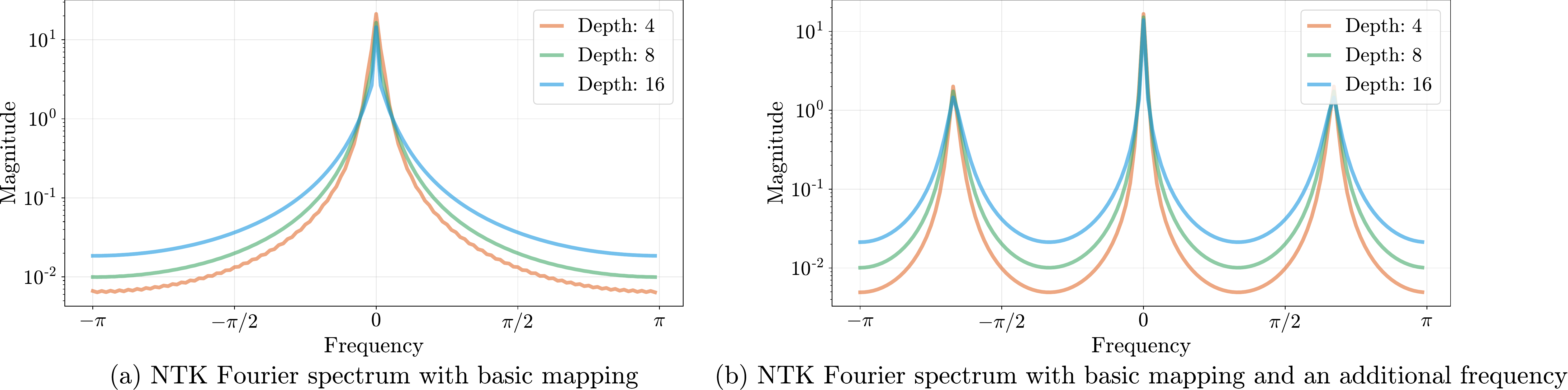}
    \caption{Each frequency included in a Fourier embedding is ``spread'' by the NTK, with deeper NTKs causing more frequency spreading. We posit that this frequency spreading is what enables an MLP with a sparse set of Fourier features to faithfully reconstruct a complex signal, which would be poorly reconstructed by either sparse Fourier feature regression or a plain coordinate-based MLP.}
    \label{fig:frequency_spreading}
\end{figure}

Again drawing on NTK theory, we understand this tradeoff as an effect of frequency ``spreading,'' as illustrated in Figure~\ref{fig:frequency_spreading}. A Fourier featurization consists of only discrete frequencies,
but when composed with the NTK, the influence of each discrete frequency ``spreads'' over its local neighborhood in the final spectrum. We find that the ``spread'' around each frequency feature increases for deeper networks. 
For an MLP to learn all of the frequency components in the target signal, its corresponding composed NTK must contain adequate power across the frequency support of the target signal. This is accomplished either by including more frequencies in the Fourier features or by spreading those frequencies through sufficient NTK depth. 

\subsection{Gradient descent does not optimize Fourier features}

\begin{figure}[h!]
    \centering
    \includegraphics[width=\textwidth]{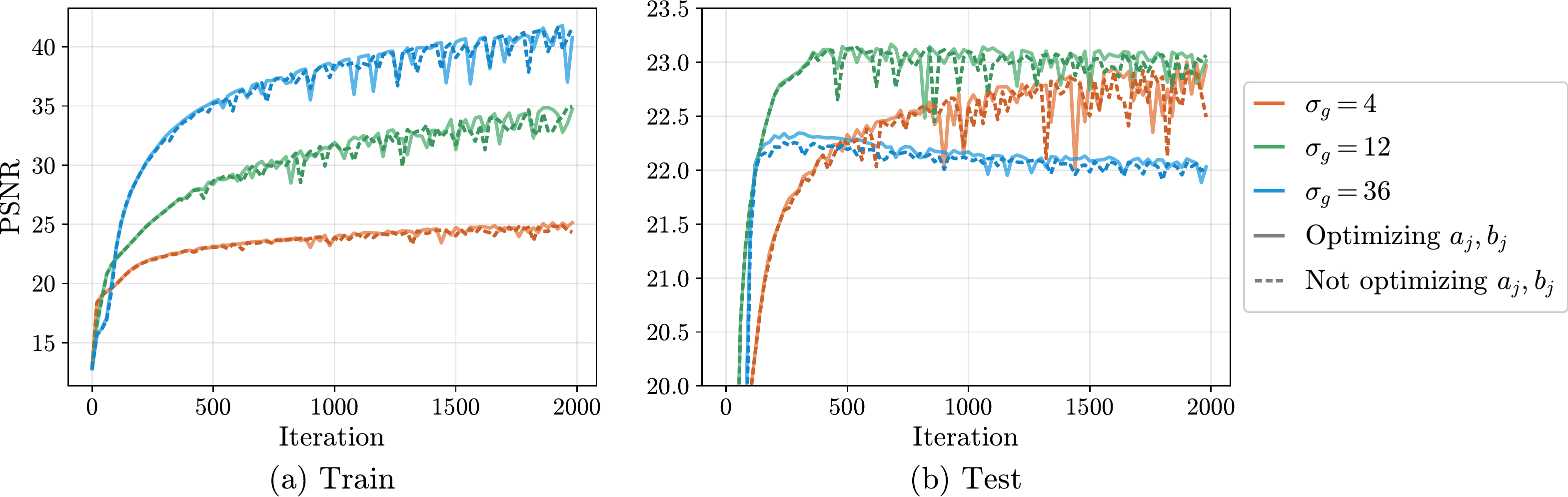}
    \caption{``Training'' the Fourier feature mapping parameters $a_j$ and $\mathbf b_j$ along with the network weights using Adam does not improve performance, as the $\mathbf b_j$ values do not deviate significantly from their initial values. We show that this holds when $\mathbf b_j$ are initialized at three different scales of Gaussian Fourier features in the case of the 2D image task ($a_j$ are always initialized as $1$).
    }
    \label{fig:no_descent}
\end{figure}

One may wonder if the Fourier feature mapping parameters $a_j$ and $\mathbf b_j$ can be optimized alongside network weights using gradient descent, which may circumvent the need for careful initialization.
We performed an experiment in which the $a_j, \mathbf b_j$ values are treated as trainable variables (along with the weights of the network) and optimize all variables with Adam to minimize training loss.
Figure~\ref{fig:no_descent} shows that jointly optimizing these parameters does not improve performance compared to leaving them fixed.

\clearpage

\subsection{Visualizing underfitting and overfitting in 2D}
\label{sec:supp_2d_scatter}

\begin{figure}[h]
    \centering
    \begin{subfigure}{.49\textwidth}
      \centering
      % include first image
      \includegraphics[width=\linewidth]{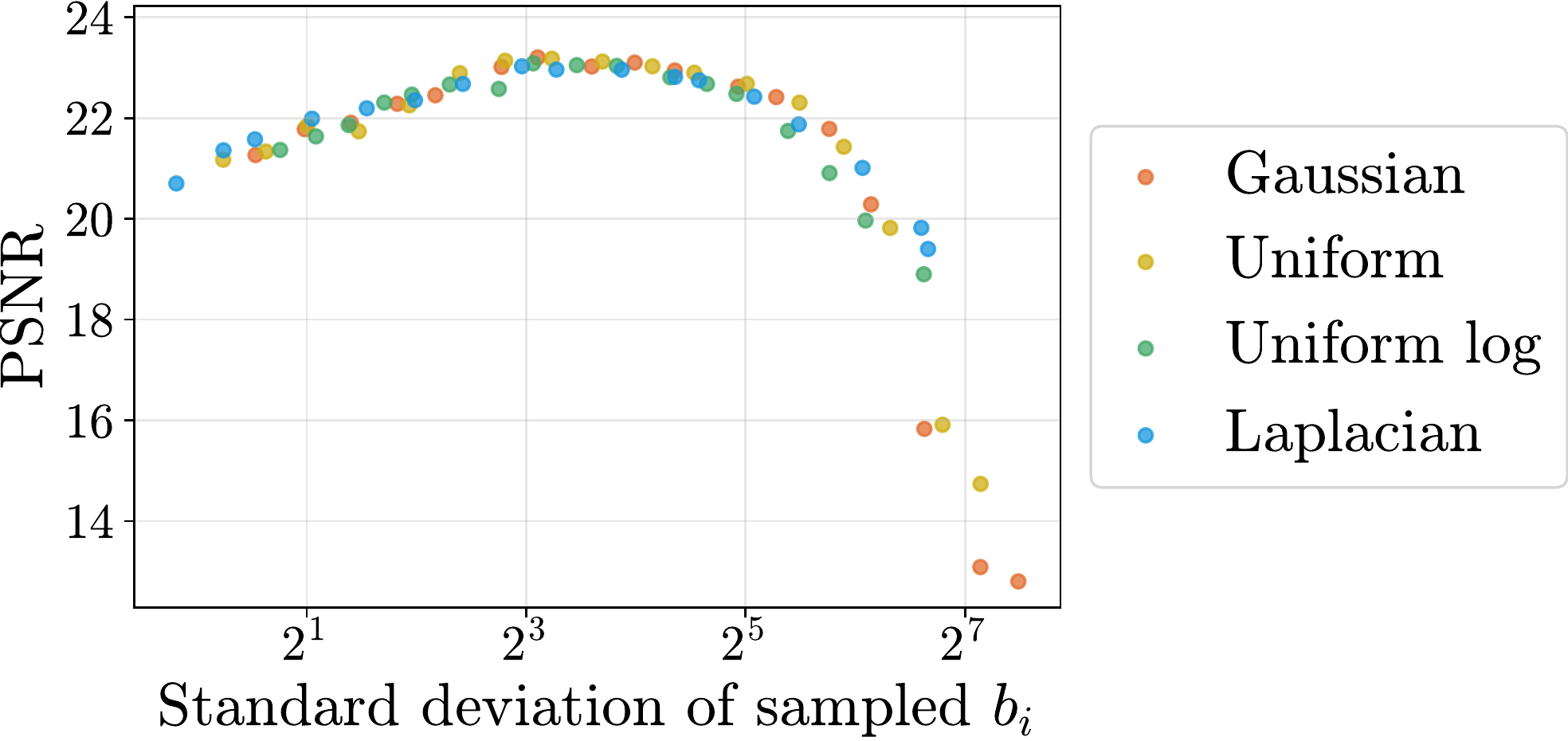}  
      \caption{Test error for 2D image task}
      \label{fig:sub-first}
    \end{subfigure}\,\,
    \begin{subfigure}{.49\textwidth}
      \centering
      % include second image
      \includegraphics[width=\linewidth]{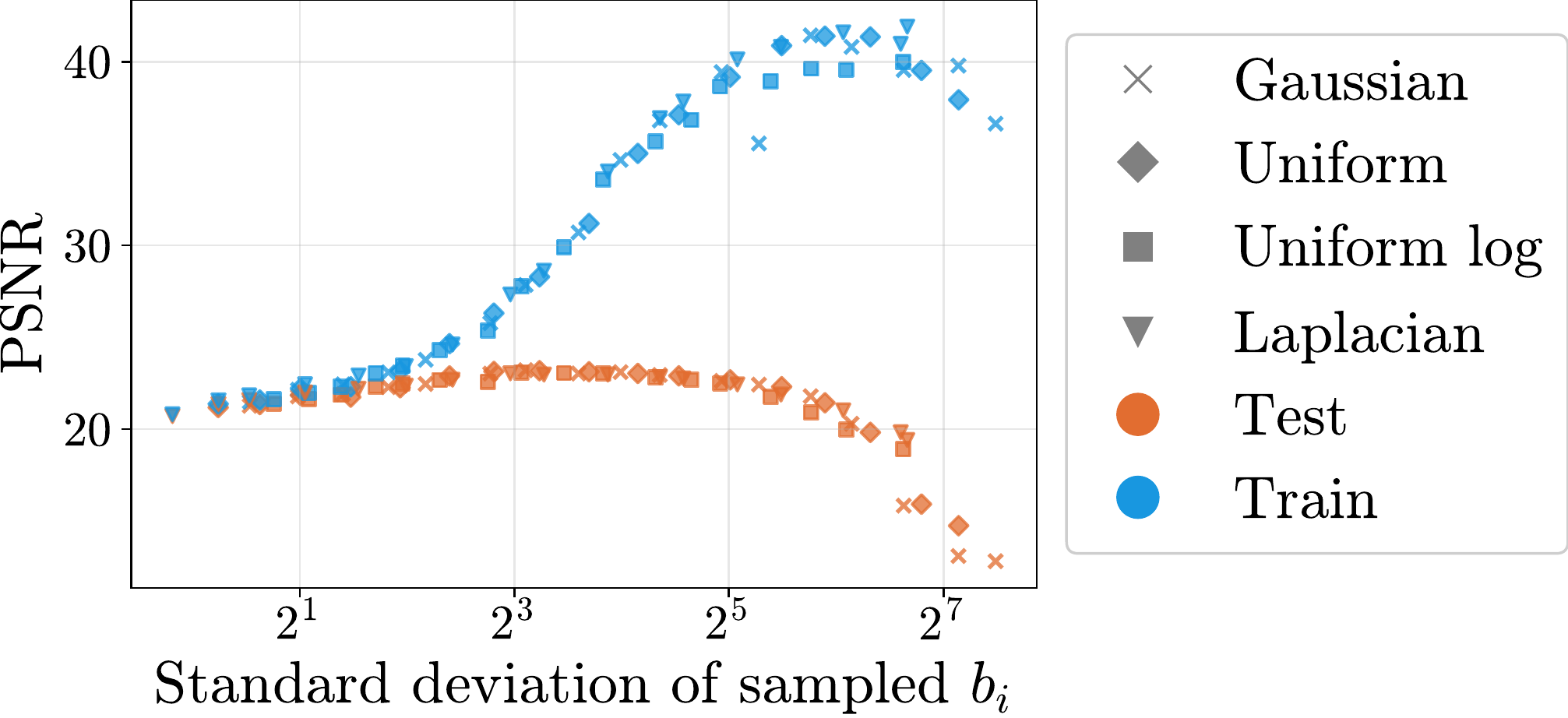}  
      \caption{Train and test error for 2D image task}
      \label{fig:sub-second}
    \end{subfigure}
    \caption{An alternate version of Figure~\ref{fig:1d_sparse} from the main text where the underlying signal is a 2D image (see 2D image task details in Section~\ref{sec:2d_image}) instead of 1D signal.
    This multi-dimensional case exhibits the same behavior as was seen in the 1D case: we see the same underfitting/overfitting pattern for four different isotropic Fourier feature distributions, and the distribution shape matters less than the scale of sampled $b_i$ values.}
    \label{fig:2d_scatter}
\end{figure}

\begin{figure}[h]
\centering

\begin{tabular}{@{}c@{\,\,}c@{\,\,}c@{\,\,}c@{\,\,}c@{}}
\includegraphics[width=\picwidth]{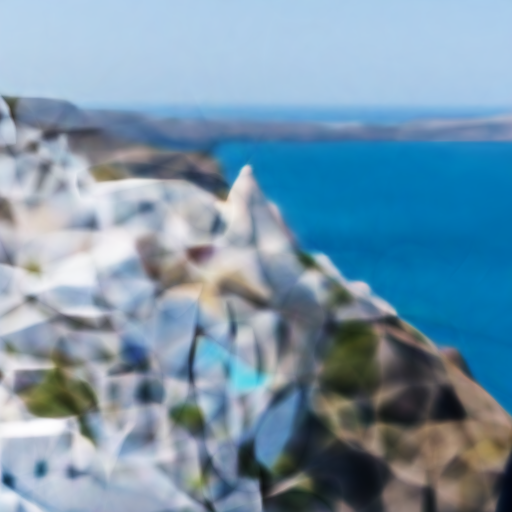} &
\includegraphics[width=\picwidth]{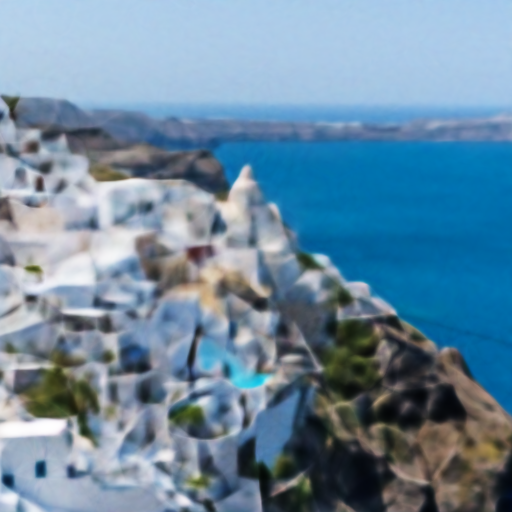} &
\includegraphics[width=\picwidth]{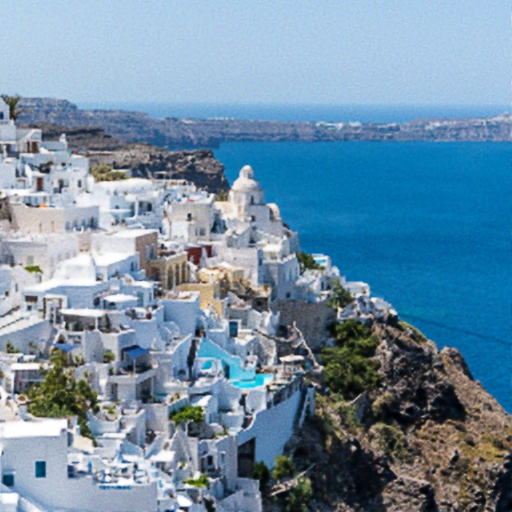} &
\includegraphics[width=\picwidth]{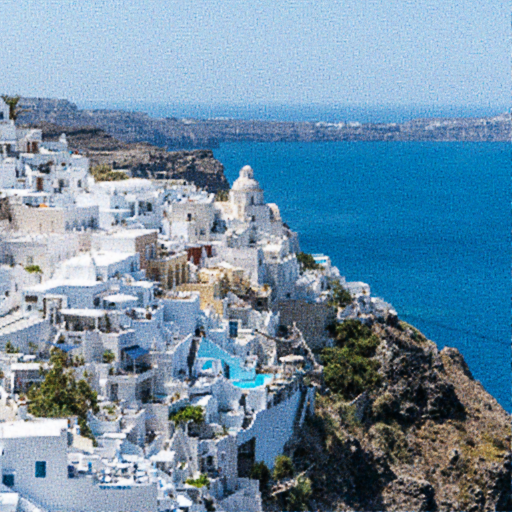} &
\includegraphics[width=\picwidth]{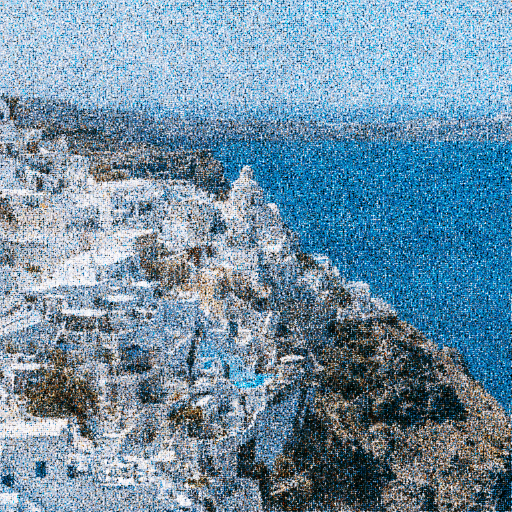}\\

\small{$\sigma=1$}
& \small{$\sigma=2$}
& \small{$\sigma=10$}
& \small{$\sigma=32$}
& \small{$\sigma=64$}
\end{tabular}
\caption{A visualization of the 2D image regression task with different Gaussian scales (corresponding to points along the curve shown in Figure~\ref{fig:2d_scatter}). 
Low values of $\sigma$ underfit, resulting in oversmoothed interpolation, and large values of $\sigma$ overfit, resulting in noisy interpolation. We find that $\sigma = 10$ performs best for our \emph{Natural} image dataset.}
\label{fig:different_scales}
\end{figure}

Figure~\ref{fig:1d_sparse} in the main text shows (in a 1D setting) that as the scale of the Fourier feature sampling distribution increases, the trained network's error traces out a curve that starts in an underfitting regime (only low frequencies are learned) and ends in an overfitting regime (the learned function includes high-frequency detail not present in the training data). In Figure~\ref{fig:2d_scatter}, we show analogous behavior for 2D image regression, demonstrating that the same phenomenon holds in a multidimensional problem.
In Figure~\ref{fig:different_scales}, we show how changing the scale for Gaussian Fourier features qualitatively affects the final result in the 2D image regression task.

\subsection{Failures of positional encoding (axis-aligned bias)}
\label{sec:axis-aligned}

\begin{figure}[h]
    \centering
    \includegraphics[width=0.75\textwidth]{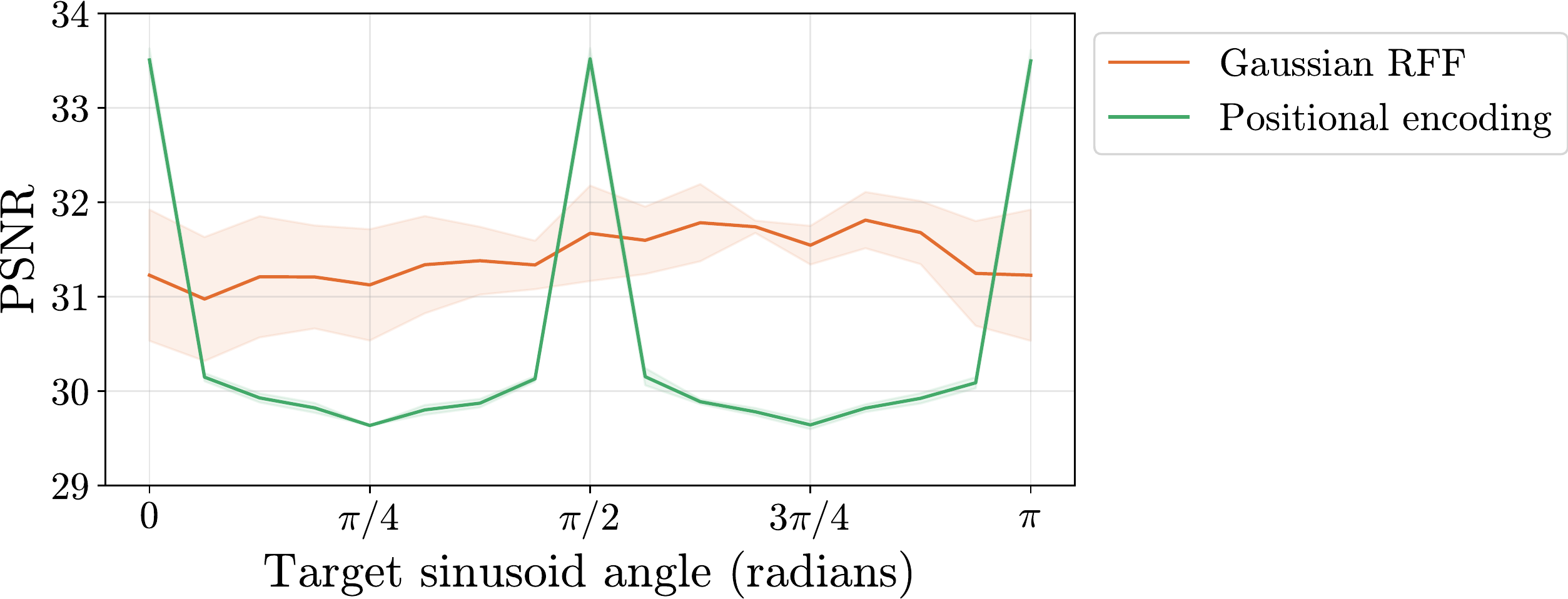}
    \caption{We train a coordinate-based MLP  to fit target 2D images consisting of simple sinusoids at different frequencies and angles. The positional encoding mapping performs well at on-axis angles and performs worse on off-axis angles, while the Gaussian RFF mapping performs similarly well across all angles (results are averaged over radii). Error bars are plotted over runs with different randomly-sampled frequencies for the Gaussian RFF mapping, while positional encoding is deterministic.}
    \label{fig:axis-aligned}
\end{figure}

Here we present a simple experiment to directly showcase the benefits of using an isotropic frequency distribution, such as Gaussian RFF, compared to the axis-aligned ``positional encoding'' used in prior work~\cite{mildenhall2020nerf,Zhong2020Reconstructing}. As discussed in the main paper, the positional encoding mapping only uses on-axis frequencies. This approach is well-suited to data that has more frequency content along the coordinate axes, but is not as effective for more natural signals.

In Figure~\ref{fig:axis-aligned}, we conduct a simple 2D image experiment where we train a coordinate-based MLP (2 layers, 256 channels) to fit target 2D sinusoid images ($512\times512$ resolution). We sample 64 such 2D sinusoid images (regularly-sampled in polar coordinates, with 16 angles and 4 radii) and train a 2D coordinate-based MLP to fit each, using the same setup as the 2D image experiments described in Section~\ref{sec:2d_image}. The isotropic Gaussian RFF mapping performs well across all angles, while the positional encoding mapping performs worse for frequencies that are not axis-aligned.

\section{Additional details for main text figures}

\subsection{Main text Figure~\ref{fig:1d_dense} (effect of feature mapping on convergence speed)}

\begin{figure}[h]
    \centering
    \includegraphics[width=\textwidth]{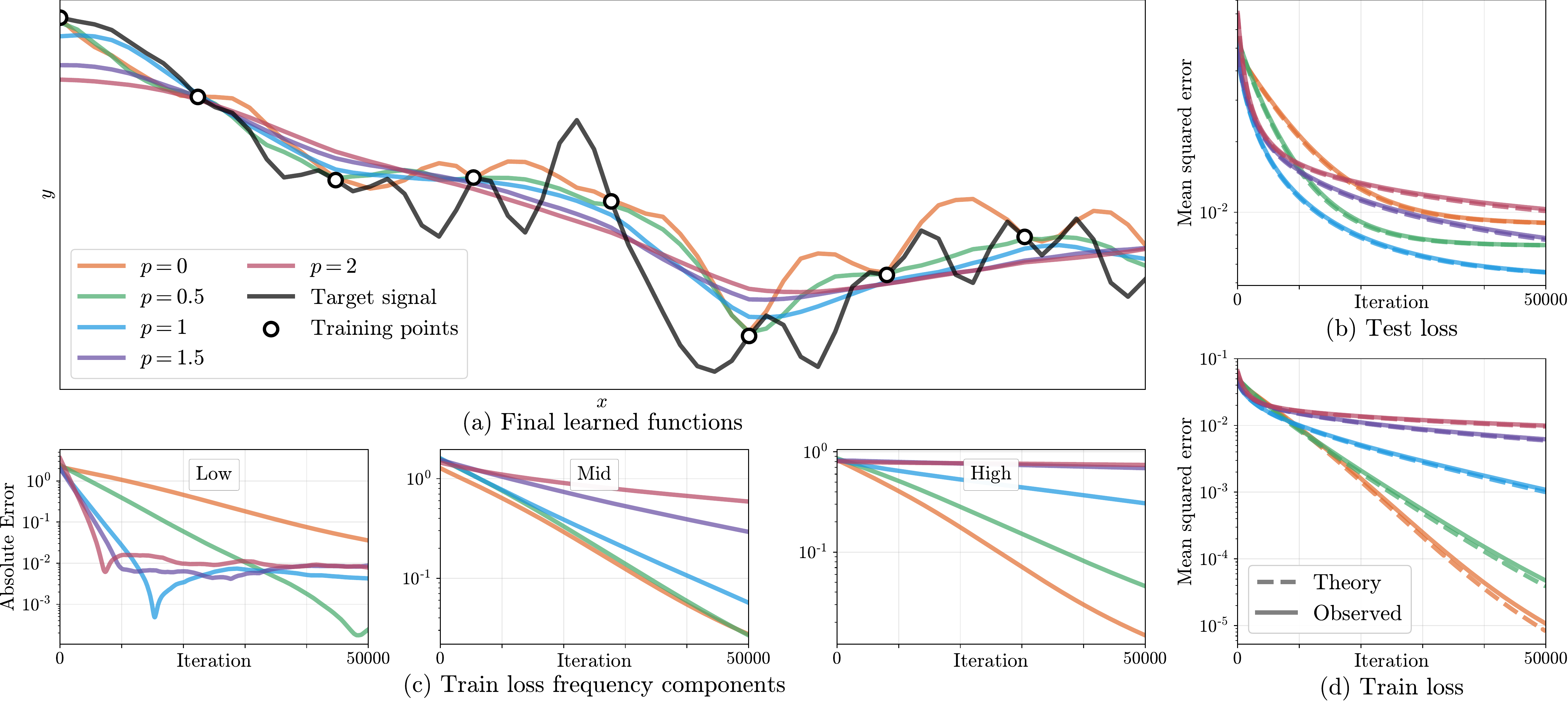}
    \caption{An extension of Figure~\ref{fig:1d_dense} from the main paper, showing more values of $p$. In (c) we see that mappings with more gradual frequency falloff (lower $p$) converge significantly faster in mid and high frequencies, resulting in faster overall training convergence (d). In (b) we see that $p=1$ achieves a lower test error than the other mappings.}
    \label{fig:more_alphas}
\end{figure}

In Figure~\ref{fig:more_alphas}, we present an alternate version of Figure 3 from the main text showing a denser sampling of $p$ values to better visualize the effect of changing Fourier feature falloff on the resulting trained network. Again, the feature mapping used here is $a_j = 1/j^p, b_j = j$ for $j=1,\ldots,\num/2$.

\subsection{Main text Figure~\ref{fig:1d_sparse} (different random feature distributions in 1D)}

\begin{figure}[h]
    \centering
    \includegraphics[width=\textwidth]{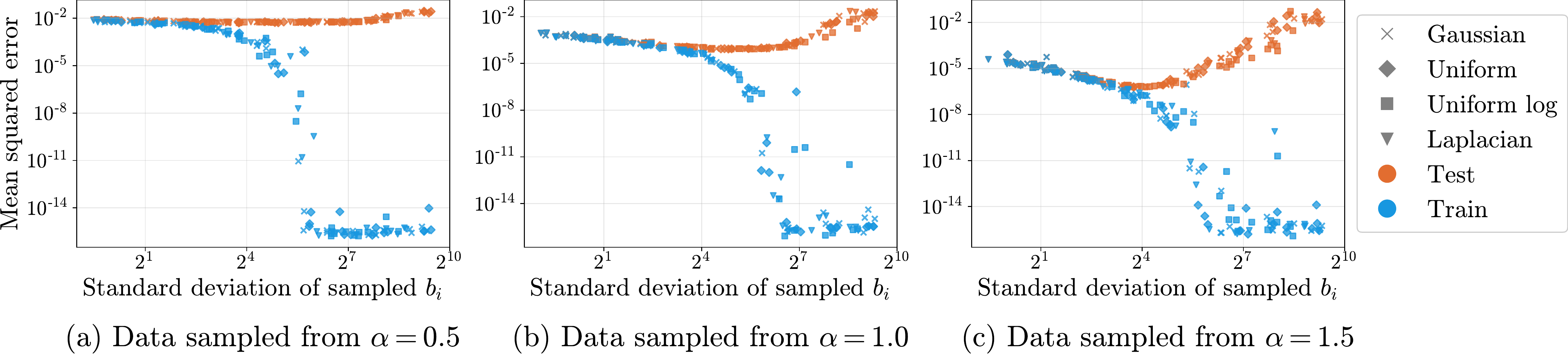}
    \caption{An alternate version of Figure~\ref{fig:1d_sparse} from the main text showing both training error and test error for a variety of different Fourier feature sampling distributions. Adding training error to the plot clearly distinguishes between the underfitting regime with low frequency $b_i$ (where train and test error are similar) versus the overfitting regime with high frequency $b_i$ (where the test error increases but training error approaches machine precision).
    }
    \label{fig:train_test}
\end{figure}

Exact details for the sampling distributions used to generate $b_j$ values for Figure~\ref{fig:1d_sparse} in the main text are shown in Table~\ref{table:distributions}. In Figure~\ref{fig:train_test}, we present an alternate version showing both train and test performance, emphasizing the underfitting/overfitting regimes created by manipulating the scale of the Fourier features.

\paragraph{Uniform log distribution} We include the \emph{Uniform log} distribution because it is the random equivalent of the ``positional encoding'' sometimes used in prior work. One observation is that the sampling for uniform-log variables ($X' = \sigma_{ul}^X$ where $X \sim \mathcal U[0,1)$) corresponds to the following CDF:
\begin{align}
    P(X' \leq x) = \frac{\log x}{\log \sigma_{ul}}, \quad\textrm{for } x \in [1, \sigma_{ul})\,,
\end{align}
which has the following PDF:
\begin{align}
    p(x) = \frac{d}{dx} P(X' \leq x)  = \frac{1}{x \log \sigma_{ul}} \,.
\end{align}
This shows that the randomized equivalent of positional encoding is sampling from a distribution proportional to a $1/f$ falloff power law.

\begin{table}[h]
    \centering
    \begin{tabular}{l|c}
         Name & Sampled $b_j$ values \\ \hline
         Gaussian    & $\sigma_g X$ for $X \sim \mathcal N(0,1)$ \\
         Uniform     & $\sigma_u X$ for $X \sim \mathcal U[0,1)$\\
         Uniform log & $\sigma_{ul}^X$ for $X \sim \mathcal U[0,1)$ \\
         Laplacian   & $\sigma_l X$ for $X \sim \mathrm{Laplace}(0,1)$ \\ \hline
         Positional Enc.   & $2^{\sigma_p X}$ for $X \in \mathrm{linspace}(0,1)$ (deterministic)
    \end{tabular}
    \vspace{2mm}
    \caption{Different distributions used for sampling frequencies, where $\sigma$ is each distribution's ``scale''.
    }
    \label{table:distributions}
\end{table}

\section{Stationary kernels}
\label{sec:stationarity}

One of the primary benefits of our Fourier feature mapping is that it results in a \emph{stationary} composed NTK function. In this section, we offer some intuition for why stationarity is desirable for our low-dimensional graphics and imaging problems. 

First, let us consider the implications of using an MLP applied directly to a low-dimensional input (without any Fourier feature mapping). In this setting, the NTK is a function of the dot product between its inputs and of their norms \cite{basri2020frequency, bietti2019inductive, bordelon2020spectrum, jacot2018neural}. 
This makes the NTK \emph{rotation}-invariant, but not \emph{translation}-invariant. For our graphics and imaging applications, we want to be able to model an object or scene equally well regardless of its location, so translation-invariance or \emph{stationarity} is a crucial property. We can then add approximate rotation invariance back by using an isotropic frequency sampling distribution.

This aligns with standard practice in signal processing, in which $k(\uvec, \vvec) = \tilde h(\uvec - \vvec) = \tilde h(\vvec - \uvec)$ (\eg the Gaussian or radial basis function kernel, or the sinc reconstruction filter kernel). This Euclidean notion of similarity based on difference vectors is better suited to the low-dimensional regime, in which we expect (and can afford) dense and nearly uniform sampling. Regression with a stationary kernel corresponds to reconstruction with a convolution filter: new predictions are sums of training points, weighted by a function of Euclidean distance. 

One of the most important features of our sinusoidal input mapping is that it translates between these two regimes. If $\uvec, \vvec \in \R^d$ for small $d$, $\gamma$ is our Fourier feature embedding function, and $k$ is a dot product kernel function, then $k(\gamma(\uvec), \gamma(\vvec)) = h(\gamma(\uvec)^\transpose \gamma(\vvec)) = \tilde h(\uvec - \vvec)$. In words, our sinusoidal input mapping transforms a dot product kernel into a stationary one, making it better suited to the low-dimensional regime.

\begin{figure}[h!]
    \centering
    \includegraphics[width=\textwidth]{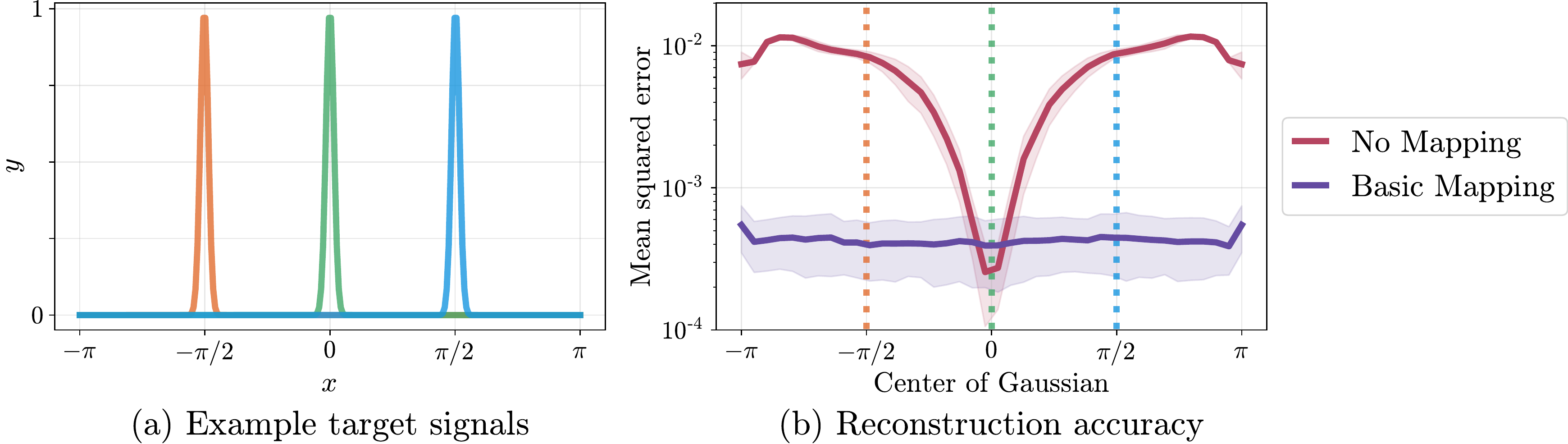}
    \caption{A plain coordinate-based MLP can learn a centered function (in this case a Gaussian density) but struggles to model shifts of the same function. Adding a basic Fourier embedding (with a single frequency) enables the MLP to fit the target function equally well regardless of shifts. The NTK corresponding to the plain MLP is based on dot products between inputs, whereas the NTK corresponding to the NTK with Fourier embedding is based on Euclidean distances between inputs, making it shift-invariant. In this experiment we train an MLP (4 layers, 256 channels, ReLU activation) for 500 iterations using the Adam~\cite{adam} optimizer with a learning rate of $10^{-4}$. We report mean and standard deviation performance over 20 random network initializations.}
    \label{fig:stationary}
\end{figure}

This effect is illustrated in a simple 1D example in Figure~\ref{fig:stationary}, which shows that the benefits of a stationary composed NTK indeed appear in the MLP setting with a basic Fourier featurization (using a single frequency). We train MLPs with and without this basic Fourier embedding to learn a set of shifted 1D Gaussian probability density functions. The plain MLP successfully fits a zero-centered function but struggles to fit shifted functions, while the MLP with basic Fourier embedding exhibits stationary behavior, with good performance regardless of shifts.

\section{Indirect supervision through a linear map}
\label{sec:linear map}

In some of the tasks we explore in this work, such as image regression or 3D shape regression, optimization is performed by minimizing a loss between the output of a network and a directly observed quantity, such as the color of a pixel or the occupancy of a voxel.
But in many graphics and imaging applications of interest, measurements are \emph{indirect}, and the loss must be computed on the output of a network after it has been processed by some physical forward model. In NeRF~\cite{mildenhall2020nerf}, measurements are taken by sampling and compositing along rays in each viewing direction. In MRI, measurements are taken along various curves through the frequency domain. In CT, measurements are integral projections of the subject at various angles, which correspond to measuring lines through the origin in the frequency domain.
Although the measurement transformation for NeRF is nonlinear (in density, although it is linear in color), those for both CT and MRI are linear. 
In this section, we extend the linearized training dynamics of Lee \etal \cite{lee2019wide} to the setting of training through a linear operator denoted by a matrix $\Linear$. This allows us to modify Eqn.~\testvaleq~ to incorporate $\Linear$, thereby demonstrating that the conclusions drawn in this work for the ``direct'' regression case also apply to the ``indirect'' case.

Our derivation closely follows Lee \etal \cite{lee2019wide}, and begins by replacing the neural network $f$ with its linearization around the initial parameters $\theta_0$:
\begin{equation}
f_t^{\lin}(\x) \triangleq f_0(\x) + \nabla _\theta f_0(\x)|_{\theta=\theta_0} \omega_t\,,
\end{equation}
where $\omega_t \triangleq \theta_t - \theta_0$ denotes the change in network parameters since initialization and $t$ denotes time in continuous-time gradient flow dynamics. Then \cite{lee2019wide} describes the dynamics of gradient flow:
\begin{align}
\dot{f}_t^\lin(\x) &= -\eta\hat{\Theta}_0(\x, \X)\nabla_{f_t^\lin(\X)} \loss\,,
\label{eq:gradflowdy}
\end{align}
where $\hat{\Theta}_t(\cdot, \cdot) = \nabla_\theta f_t(\cdot) \nabla_\theta f_t(\cdot)^\transpose $ is the NTK matrix at time $t$ ($\hat{\Theta}_t$ is shorthand for $\hat{\Theta}_t(\X, \X)$) and $\loss$ is the training loss.
At this point, we depart slightly from the analysis of \cite{lee2019wide}: instead of $\loss = \sum_{(\x, y) \in \D} \ell(f_t^\lin(\x), y)$ we have $\loss = \frac{1}{2}\norm{\Linear(f_t^\lin(\X) - \Y)}_2^2$, where $\Y$ denotes the vector of training labels. The gradient of the loss is then
\begin{align}
    \nabla_{f_t^\lin(\X)} \loss &= \nabla_{f_t^\lin(\X)} \frac{1}{2}\norm{\Linear \left(f_t^\lin(\X) - \Y \right)}_2^2\, \\
    &= \Linear^\transpose \Linear \left( f_t^\lin(\X) - \Y \right)\,.
\end{align}
Substituting this into the gradient flow dynamics of Eqn.~\ref{eq:gradflowdy} gives us:
\begin{align}
\dot{f}_t^\lin(\x) &= -\eta\hat{\Theta}_0(\x, \X)\Linear^\transpose \Linear\left(f_t^\lin(\X) - \Y\right)\,,
\end{align}
with corresponding solution:
\begin{align}
f_t^\lin(\X) &= \left( \Eye-e^{-\eta \hat \Theta_0 \Linear^\transpose \Linear t} \right)\Y + e^{-\eta\hat\Theta_0 \Linear^\transpose \Linear t} f_0(\X)\,.
\end{align}
Finally, again following \cite{lee2019wide}, we can decompose $f_t^\lin(\x) = \mu_t(\x) + \gamma_t(\x)$ at any test point $\x$, where 
\begin{align}
\mu_t(\x) &= \hat \Theta_0(\x, \X)\hat \Theta_0^{-1}\left(\Eye-e^{-\eta \hat \Theta_0 \Linear^\transpose \Linear t}\right)\Y\,,\\
\gamma_t(\x) &= f_0(\x) - \hat \Theta_0(\x, \X)\hat \Theta_0^{-1}\left(\Eye-e^{-\eta \hat \Theta_0 \Linear^\transpose \Linear t}\right)f_0(\X)\,.
\end{align}
Assuming our initialization is small, \ie $f_0(\x) \approx 0~\forall \x$, we can write our approximate linearized network output as:
\begin{equation}
    f_t^\lin(\x) \approx \hat \Theta_0(\x, \X)\hat \Theta_0^{-1}\left(\Eye-e^{-\eta \hat \Theta_0 \Linear^\transpose \Linear t}\right)\Y\,.
\end{equation}

In our previous analysis, we work instead with the expected or infinite-width NTK matrix $\Kernel$, which is fixed throughout training. Using this notation, we have
\begin{equation}
\hat{\mathbf{y}}^{(t)} \approx f_t^\lin(\arbitrary{\Data}) \approx \arbitrary{\Kernel} \Kernel^{-1} \left(\Eye - e^{-\eta \Kernel \Linear^\transpose \Linear t}\right)\Y\,.
\label{eqn:a_testvals}
\end{equation}

This is nearly identical to Eqn.~\testvaleq in the main paper, except that the convergence is governed by the spectrum of $\Kernel\Linear^\transpose  \Linear$ rather than $\Kernel$ alone. If $\Linear$ is unitary, such as the Fourier transform matrix used in (densely sampled) MRI, then training should behave exactly as if we were training on direct measurements. However, if $\Linear$ is not full rank, then training will only affect the components with nonzero eigenvalues in $\Kernel\Linear^\transpose  \Linear$. In this more common scenario, we want to design a kernel that will provide large eigenvalues in the components that $\Linear$ can represent, so that the learnable components will converge quickly, and provide reasonable priors for the components we cannot learn.

In our two tasks that supervise through a linear map, CT and MRI, the $\Linear^\transpose  \Linear$ has a structure that illuminates how the linear map interacts with the composed NTK.
The $\Linear^\transpose  \Linear$ matrices for both these tasks are diagonalizable by the DFT matrix, where the diagonal entries are simply the number of times the corresponding frequency is measured by the MRI or CT sampling patterns. This follows from the fact that CT and MRI measurements can both be formulated as Fourier space sampling: CT samples rotated slices in Fourier space through the origin~\cite{bracewell} and MRI samples operator-chosen Fourier trajectories. This means that frequencies not observed by the MRI or CT sampling patterns will never be supervised during training. Therefore, it is crucial to choose a Fourier feature mapping that results in a composed NTK with a good prior on these frequencies.

\section{Task details}

\label{sec:supp_tasks}

We present additional details for each task from Section~\ref{sec:experiments} in the main text, including training parameters, forward models, datasets, etc. All experiments are implemented using JAX~\cite{jax2018github} and trained on a single K80 or RTX2080Ti GPU. Training a single MLP took between 10 seconds (for the 2D image task) and 30 minutes (for the inverse rendering task).

\subsection{2D image} %%%%%%%%%%%%%%%%%%%%%%%
\label{sec:2d_image}

\begin{figure}[h!]
\centering

\begin{tabular}{@{\,\,}c@{\,\,}c@{\,\,}c@{\,\,}c@{\,\,}c@{\,\,}}
\includegraphics[width=\picwidth]{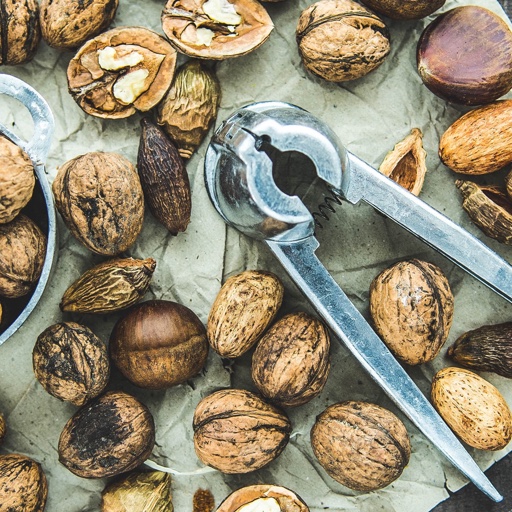} &
\includegraphics[width=\picwidth]{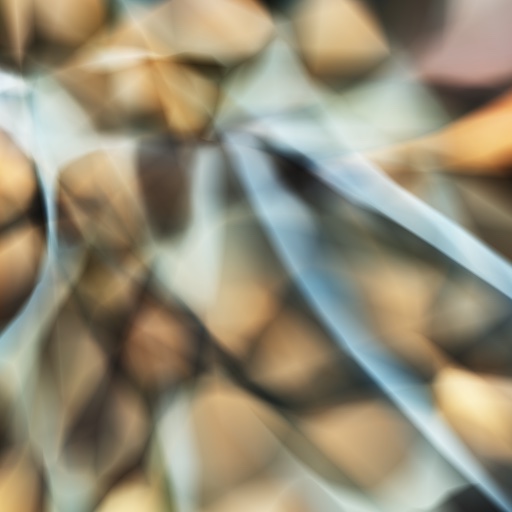} &
\includegraphics[width=\picwidth]{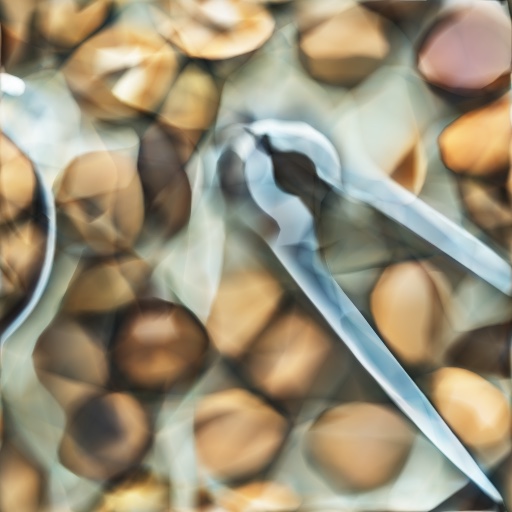} &
\includegraphics[width=\picwidth]{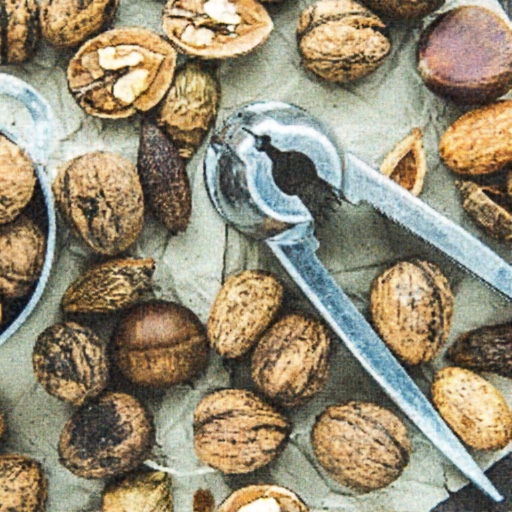} &
\includegraphics[width=\picwidth]{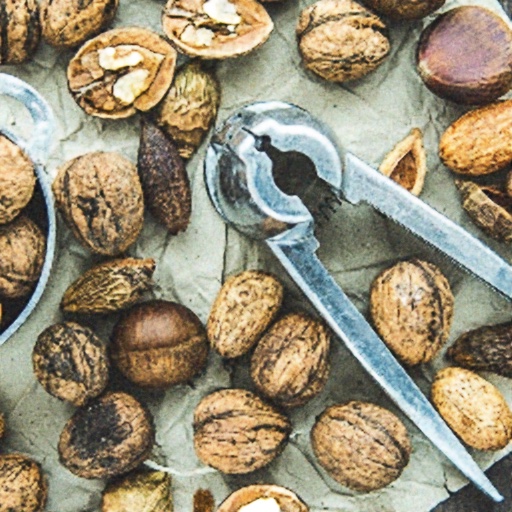}\\

\includegraphics[width=\picwidth]{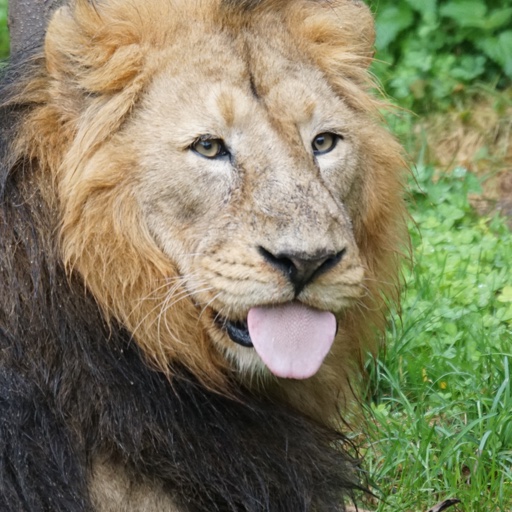} &
\includegraphics[width=\picwidth]{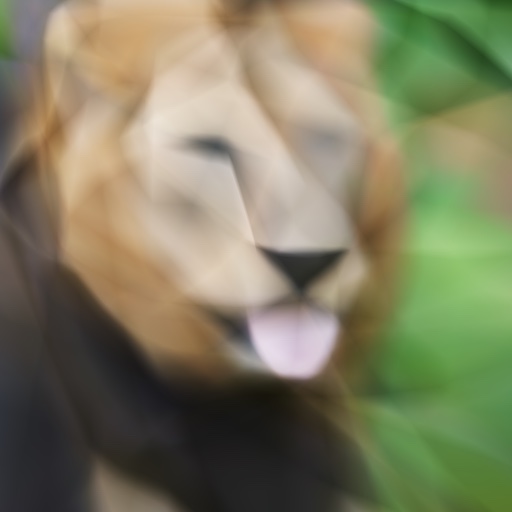} &
\includegraphics[width=\picwidth]{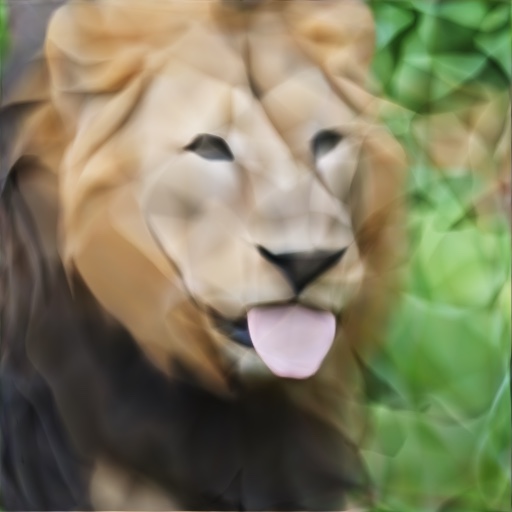} &
\includegraphics[width=\picwidth]{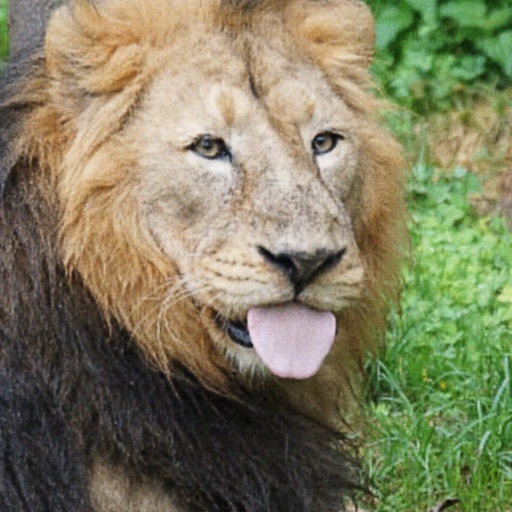} &
\includegraphics[width=\picwidth]{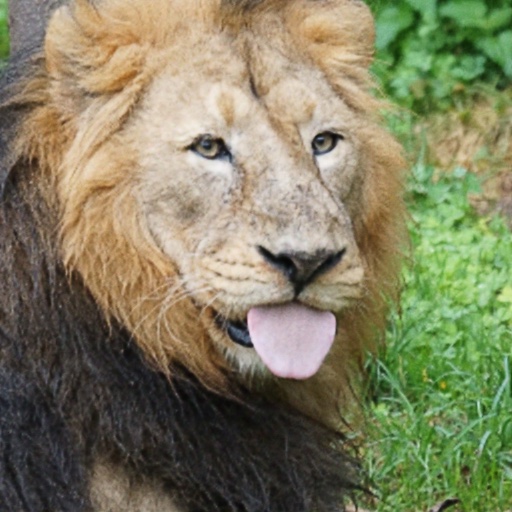}\\

\includegraphics[width=\picwidth]{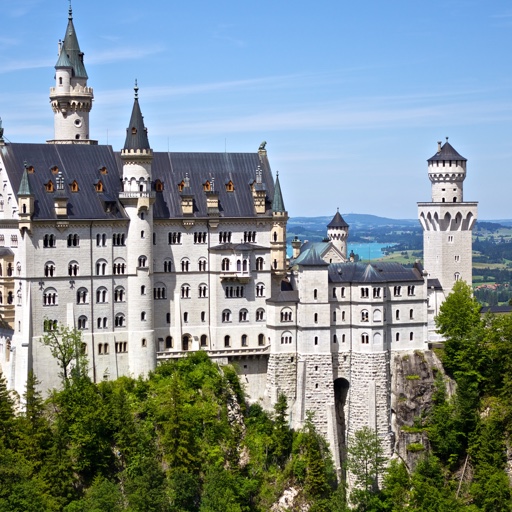} &
\includegraphics[width=\picwidth]{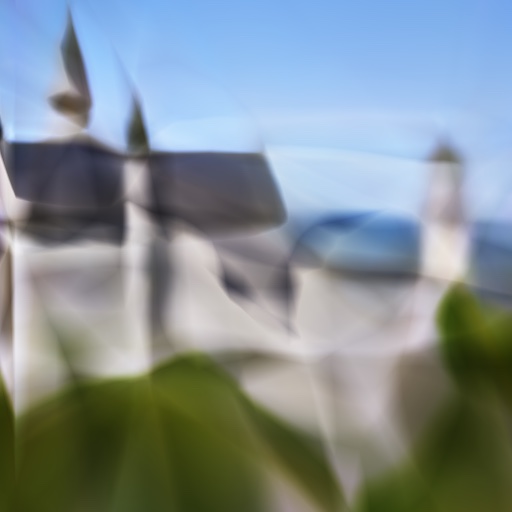} &
\includegraphics[width=\picwidth]{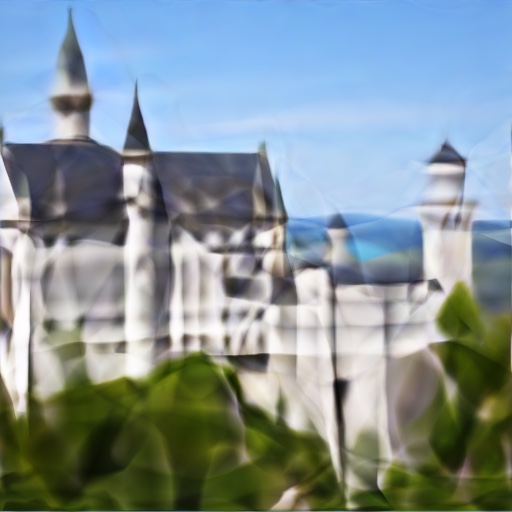} &
\includegraphics[width=\picwidth]{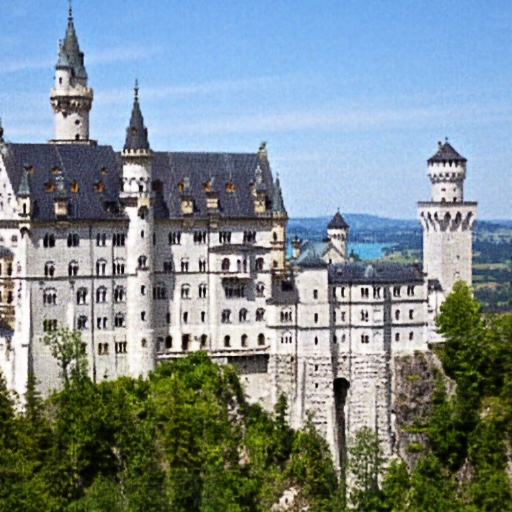} &
\includegraphics[width=\picwidth]{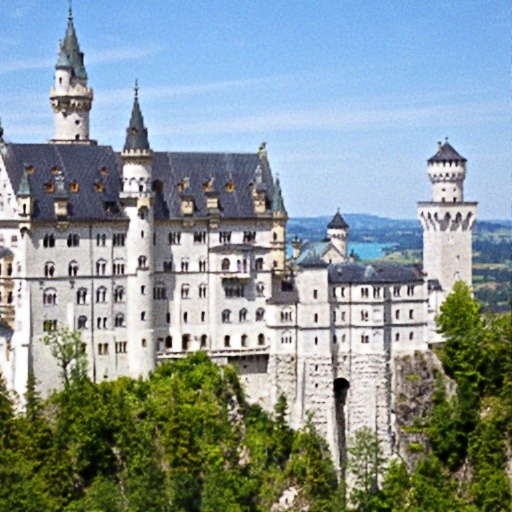}\\

\includegraphics[width=\picwidth]{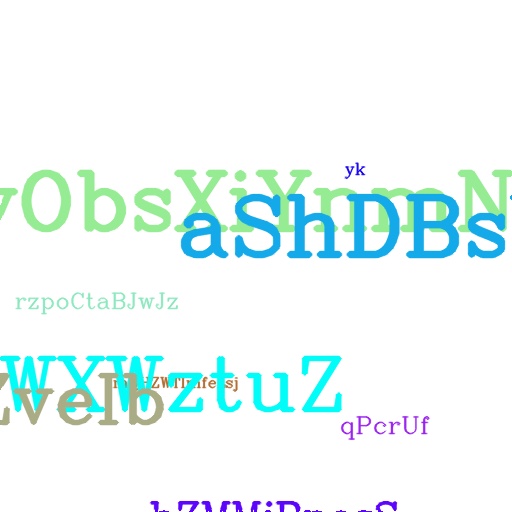} &
\includegraphics[width=\picwidth]{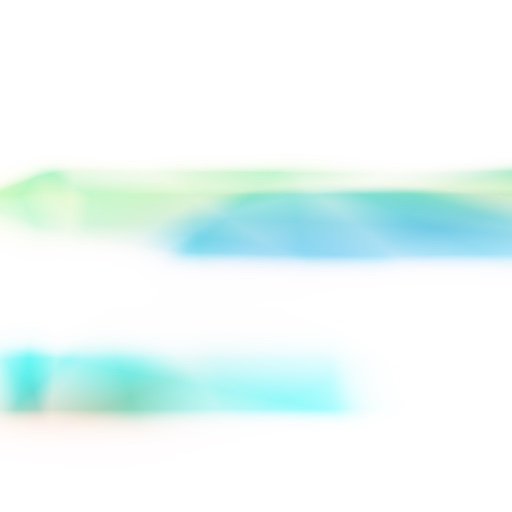} &
\includegraphics[width=\picwidth]{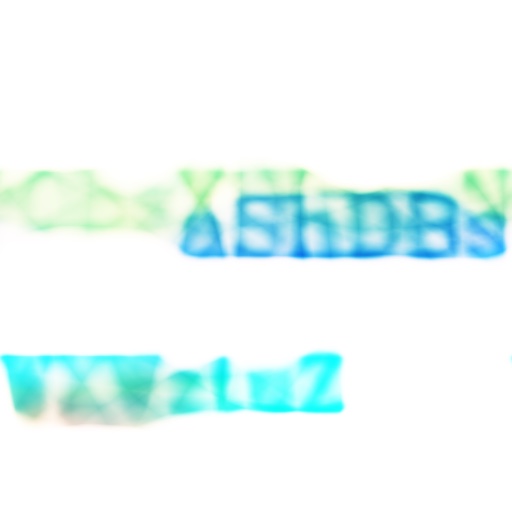} &
\includegraphics[width=\picwidth]{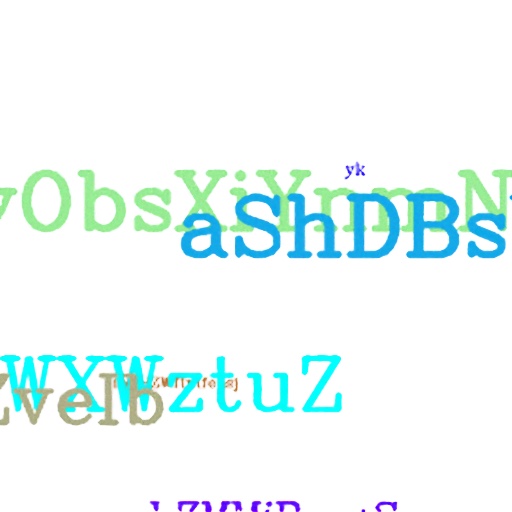} &
\includegraphics[width=\picwidth]{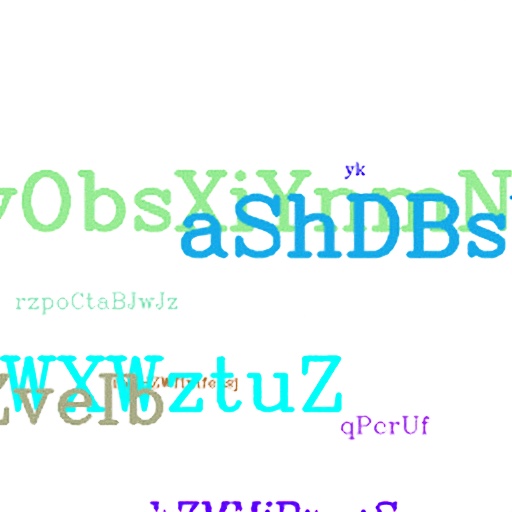}\\

\includegraphics[width=\picwidth]{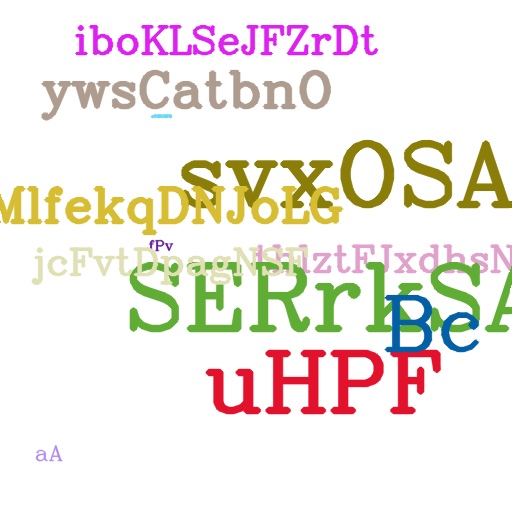} &
\includegraphics[width=\picwidth]{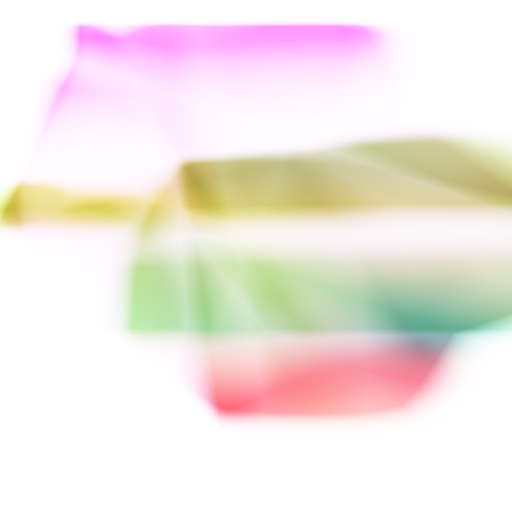} &
\includegraphics[width=\picwidth]{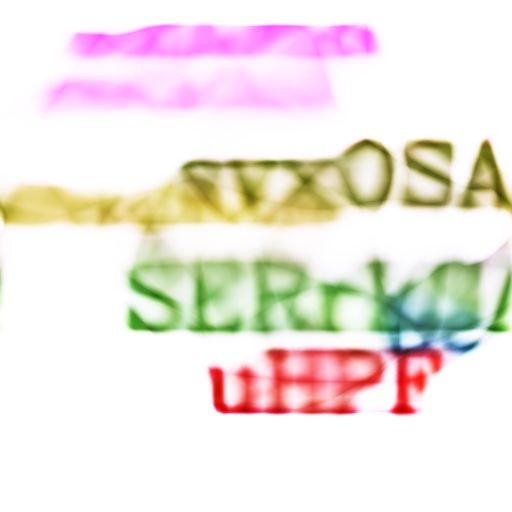} &
\includegraphics[width=\picwidth]{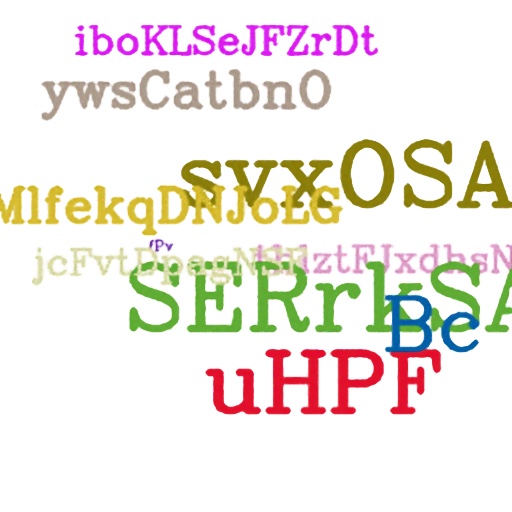} &
\includegraphics[width=\picwidth]{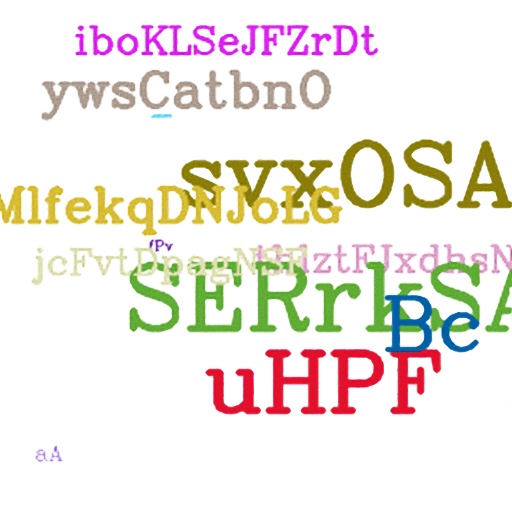}\\

\small{(a) Ground Truth}
& \small{(b) No mapping}
& \small{(c) Basic}
& \small{(d) Positional enc.}
& \small{(e) Gaussian}
\end{tabular}
\caption{Additional results for the 2D image regression task, for three images from our \emph{Natural} dataset (top) and two images from our \emph{Text} dataset (bottom).}
\label{fig:more_2d_images}
\end{figure}

The 2D image regression tasks presented in the main text all use $512\times512$ resolution images. A subsampled grid of $256\times256$ pixels is used as training data, and an offset grid of $256 \times 256$ pixels is used for testing. We use two image datasets: \emph{Natural} and \emph{Text}, each consisting of 32 images. The \emph{Natural} images are generated by taking center crops of randomly sampled images from the Div2K dataset~\cite{div2k_Agustsson_2017}. The \emph{Text} images are generated by placing random strings of text with random sizes and colors on a white background (examples can be seen in Figure~\ref{fig:more_2d_images}). For each dataset we perform a hyperparameter sweep over feature mapping scales on 16 images. We find that scales $\sigma_g=10$ and $\sigma_p=6$ work best for the \emph{Natural} dataset and $\sigma_g=14$ and $\sigma_p=5$ work best for the \emph{Text} dataset (see Table~\ref{table:distributions} for mapping definitions). In Table~\ref{table:2d_image_results}, we report model performance using the optimal mapping scale on the remaining 16 images.

\begin{table}[h]
\centering
\begin{tabular}{l|cc}
            & Natural & Text \\ \hline
           No mapping      & $19.32 \pm 2.48$ & $18.40 \pm 2.23$ \\
           Basic           & $21.71 \pm 2.71$ & $20.48 \pm 1.96$ \\
           Positional enc. & $24.95 \pm 3.72$ & $27.57 \pm 3.07$ \\
           Gaussian        & $\mathbf{25.57 \pm 4.19}$ & $\mathbf{30.47 \pm 2.11}$
\end{tabular}
\vspace{2mm}
\caption{2D image results (mean $\pm$ standard deviation of PSNR)}
\label{table:2d_image_results}
\end{table}

Each model (MLP with 4 layers, 256 channels, ReLU activation, sigmoid output) is trained for 2000 iterations using the Adam~\cite{adam} optimizer with default settings ($\beta_1=0.9$, $\beta_2=0.999$, $\epsilon=10^{-8}$). Learning rates are manually tuned for each dataset and method. For \emph{Natural} images a learning rate of $10^{-3}$ is used for the Gaussian RFF and the positional encoding, and a learning rate of $10^{-2}$ is used for the basic mapping and ``no mapping'' methods. For the \emph{Text} images a learning rate of $10^{-3}$ is used for all methods.

\subsection{3D shape} %%%%%%%%%%%%%%%%%%%%%%%

\begin{figure}[p]
\centering

\begin{tabular}{@{}c@{\,\,}c@{\,\,}c@{\,\,}c@{\,\,}c@{}}
\includegraphics[width=\picwidth]{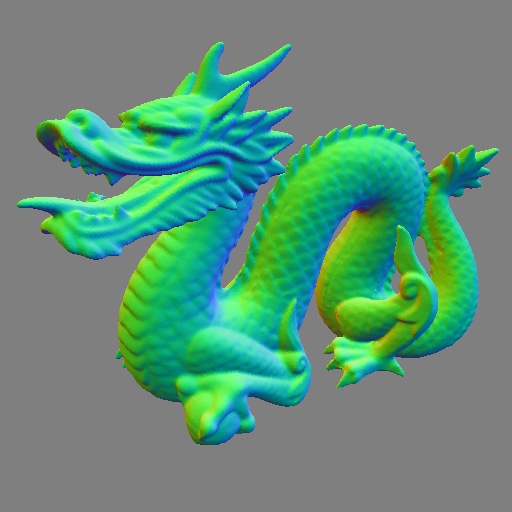} &
\includegraphics[width=\picwidth]{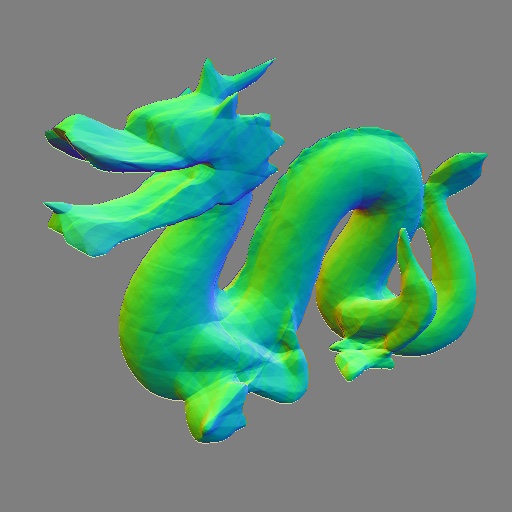} &
\includegraphics[width=\picwidth]{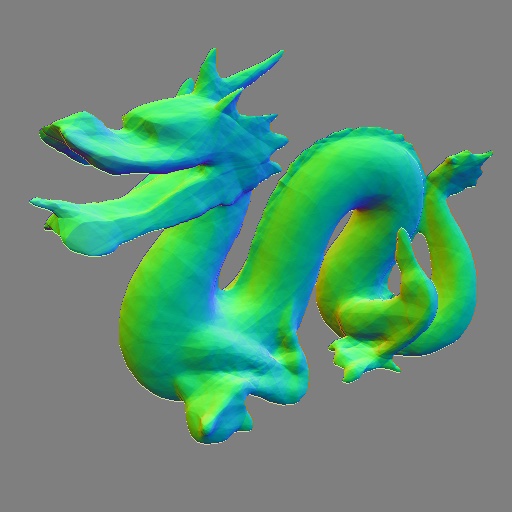} &
\includegraphics[width=\picwidth]{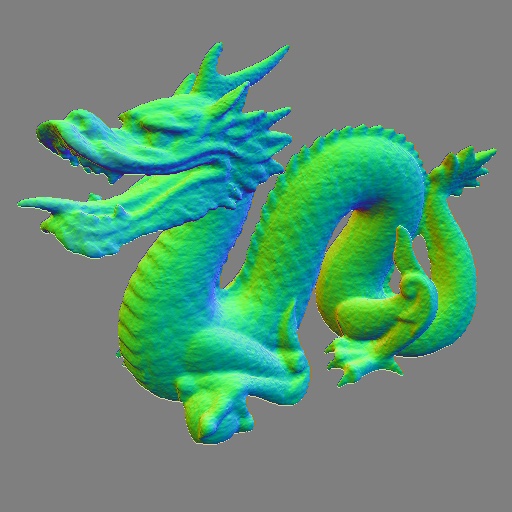} &
\includegraphics[width=\picwidth]{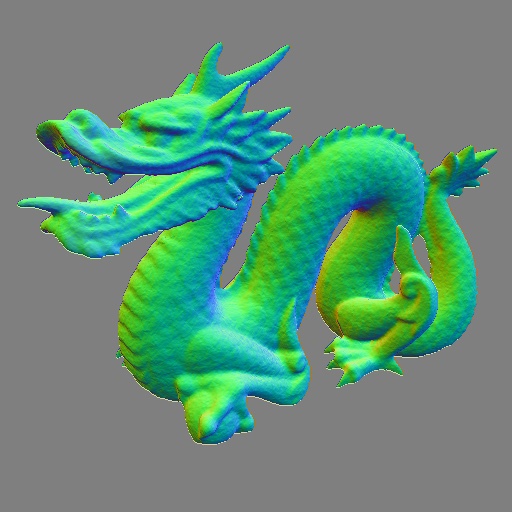}\\

\includegraphics[width=\picwidth]{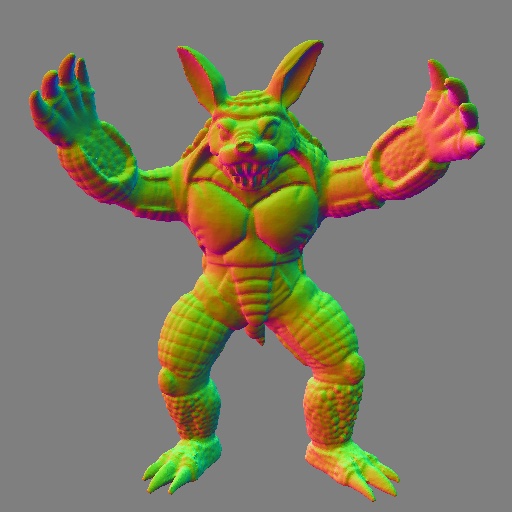} &
\includegraphics[width=\picwidth]{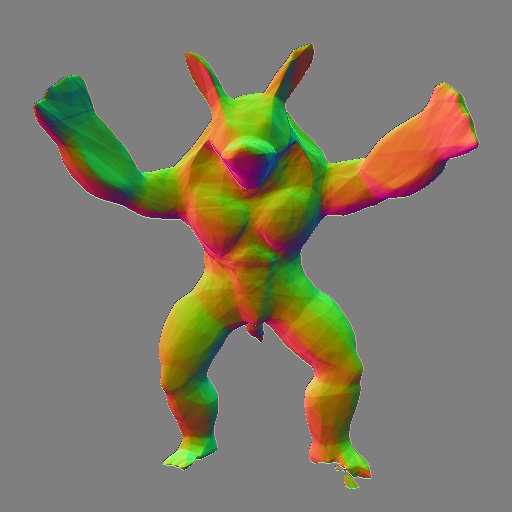} &
\includegraphics[width=\picwidth]{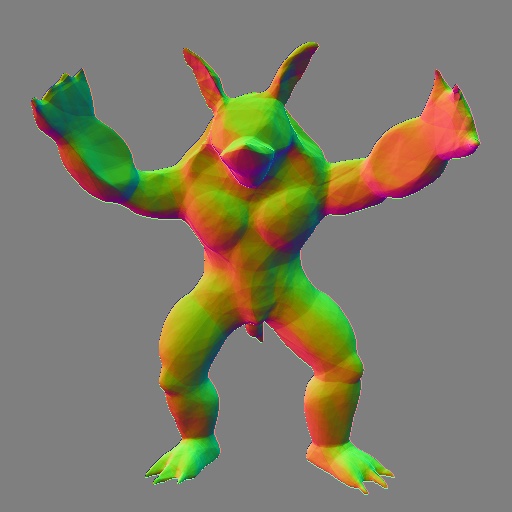} &
\includegraphics[width=\picwidth]{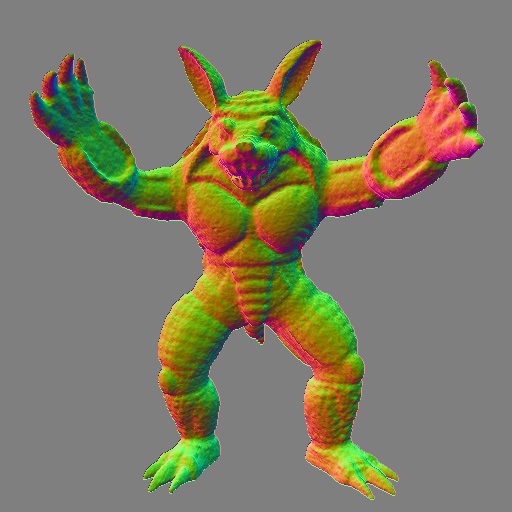} &
\includegraphics[width=\picwidth]{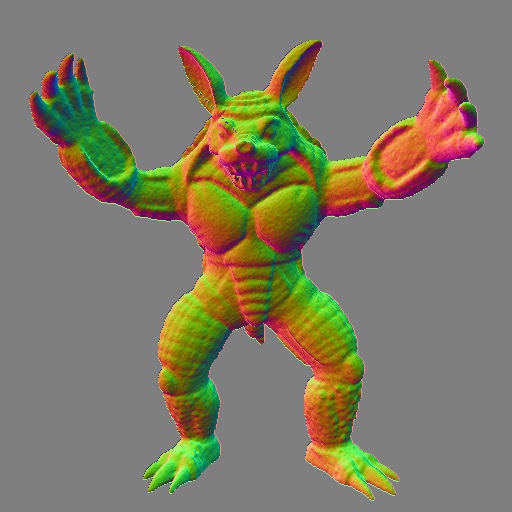}\\

\includegraphics[width=\picwidth]{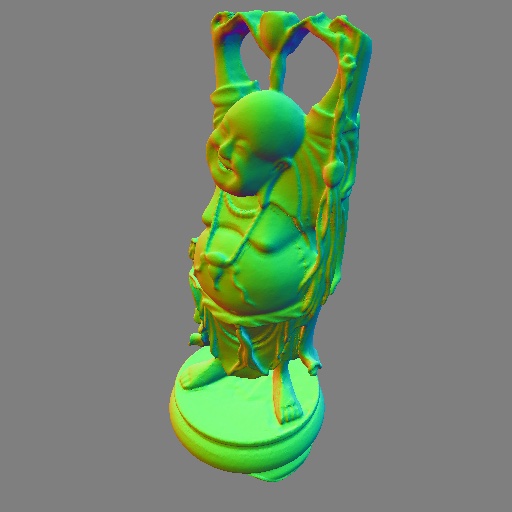} &
\includegraphics[width=\picwidth]{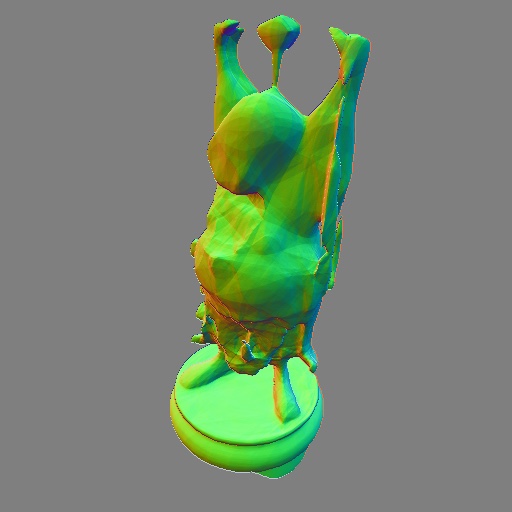} &
\includegraphics[width=\picwidth]{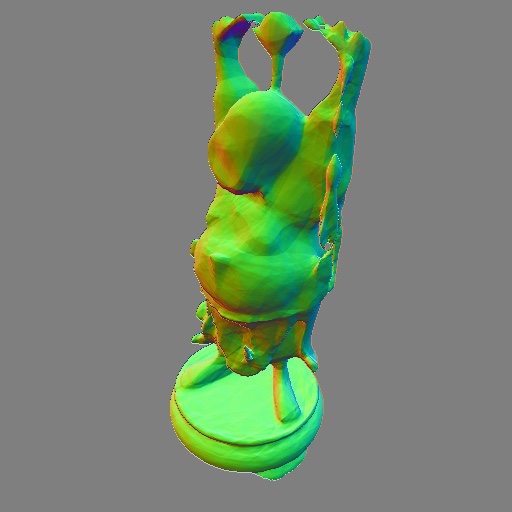} &
\includegraphics[width=\picwidth]{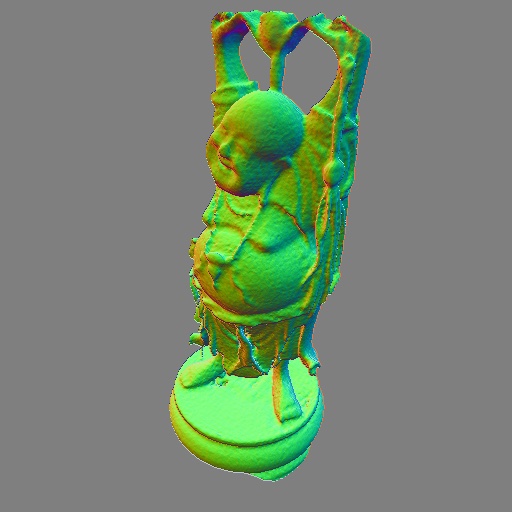} &
\includegraphics[width=\picwidth]{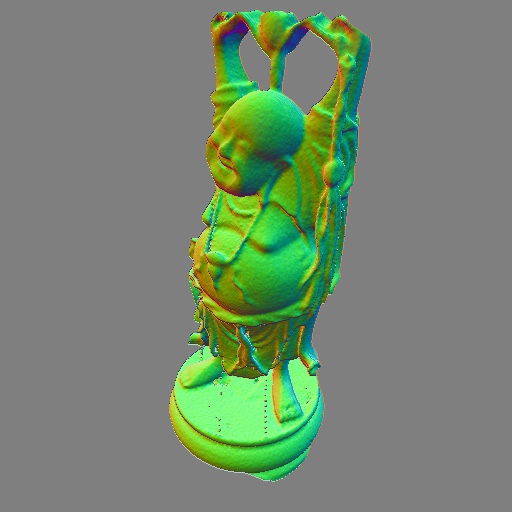}\\

\includegraphics[width=\picwidth]{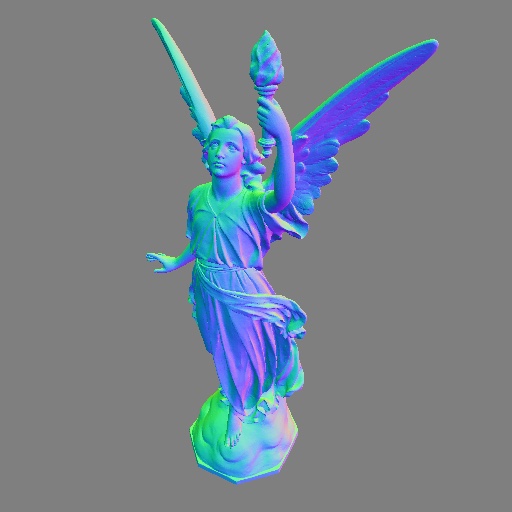} &
\includegraphics[width=\picwidth]{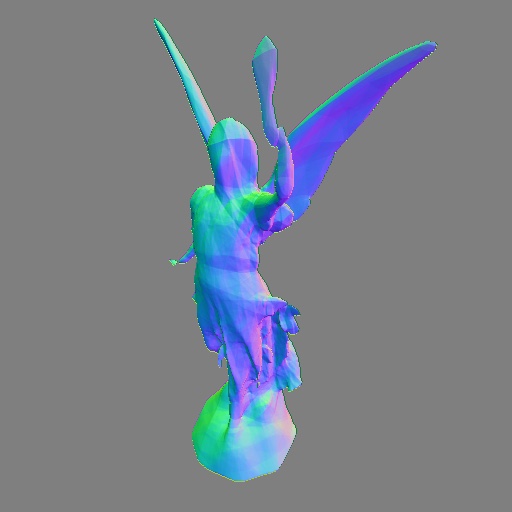} &
\includegraphics[width=\picwidth]{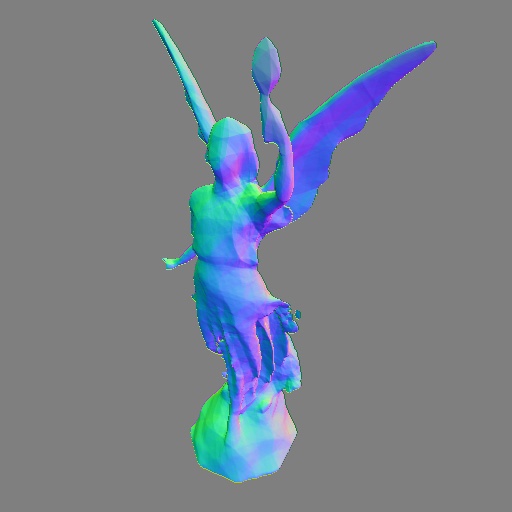} &
\includegraphics[width=\picwidth]{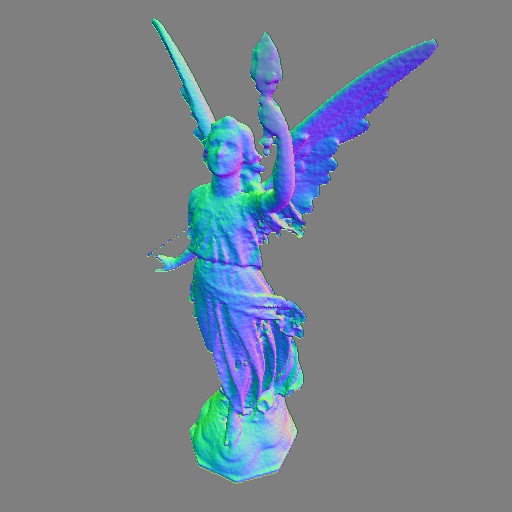} &
\includegraphics[width=\picwidth]{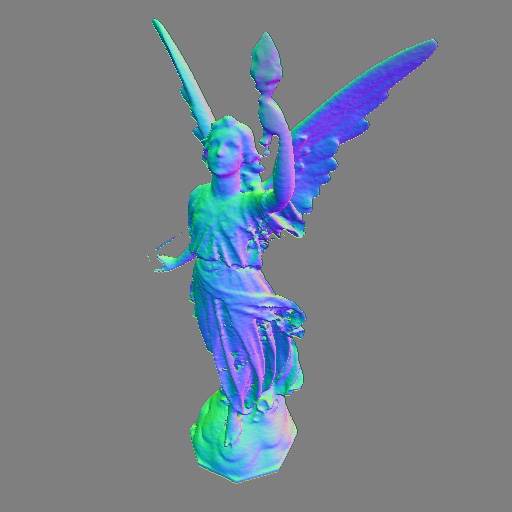}\\

\small{(a) Ground Truth}
& \small{(b) No mapping}
& \small{(c) Basic}
& \small{(d) Positional enc.}
& \small{(e) Gaussian}
\end{tabular}
\caption{Additional results for the 3D shape occupancy task~\cite{occupancynet}.}
\label{fig:more_3d_shapes}
\end{figure}

We evaluate the 3D shape regression task (similar to Occupancy Networks~\cite{occupancynet}) on four complex triangle meshes commonly used in computer graphics applications (\emph{Dragon}, \emph{Armadillo}, \emph{Buddha}, and \emph{Lucy}, shown in Figure~\ref{fig:more_3d_shapes}), each containing hundreds of thousands of vertices. We train one coordinate-based MLP network to represent a single mesh rather than trying to generalize one network to encode multiple objects, since our goal is to demonstrate that a network with no mapping or the low frequency ``basic'' mapping cannot accurately represent even a \emph{single} shape, let alone a whole class of objects.

We use a network with 8 layers of 256 channels each and a ReLU nonlinearity between each layer. We apply a sigmoid activation to the output. Our batch size is $32^3$ points, and we use the Adam optimizer~\cite{adam} with a learning rate starting at $5 \times 10^{-4}$ and exponentially decaying by a factor of $0.01$ over the course of 10000 total training iterations. At each training iteration, we sample a batch of 3D points uniformly at random from the bounding box of the mesh, and then calculate ground truth labels (using the point-in-mesh method implemented in the Trimesh library~\cite{trimesh}, which relies on the Embree kernel for acceleration~\cite{embree}). We use cross-entropy loss to train the network to match these classification labels (0 for points outside the mesh, 1 for points inside). 

The meshes are scaled to fit inside the unit cube $[0,1]^3$ such that the centroid of the mesh is $(0.5, 0.5, 0.5)$. We use the \emph{Lucy} statue mesh as a validation object to find optimal scale values for the positional encoding and Gaussian feature mapping. As described in the caption for Table~\ref{table:3d_shape_results}, we calculate error on both a uniformly random test set and a test set that is close to the mesh surface (randomly chosen mesh vertices that have been perturbed by a random Gaussian vector with standard deviation $0.01$) in order to illustrate that Fourier feature mappings provide a large benefit in resolving fine surface details. Both test sets have $64^3$ points.

\begin{table}[h!]
\centering
\begin{tabular}{l|cc}
            & Uniform points & Boundary points \\  \hline
            No mapping       & $0.959 \pm 0.006$ & $0.864 \pm 0.014$ \\
            Basic            & $0.966 \pm 0.007$ & $0.892 \pm 0.017$ \\
            Positional enc.  & $0.987 \pm 0.005$ & $0.960 \pm 0.011$ \\
            Gaussian         & $\mathbf{0.988 \pm 0.007}$ & $\mathbf{0.973 \pm 0.010}$ \\
\end{tabular}
\vspace{2mm}
\caption{3D shape results (mean $\pm$ standard deviation of intersection-over-union). \emph{Uniform points} is an ``easy'' test set where points are sampled uniformly at random from the bounding box of the ground truth mesh, while \emph{Boundary points} is a ``hard'' test set where points are sampled near the boundary of the ground truth mesh.}
\label{table:3d_shape_results}
\end{table}

In Figure~\ref{fig:more_3d_shapes}, we visualize additional results on all four meshes mentioned above (including the validation mesh \emph{Lucy}). We render normal maps, which are computed by taking the cross product of the numerical horizontal and vertical derivatives of the depth map. The original depth map is generated by intersecting camera rays with the first $0.5$ isosurface of the network. We select the Fourier feature scales for (d) and (e) by doing a hyperparameter search based on validation loss for the \emph{Lucy} mesh in the last row and report test loss over the other three meshes (Table~\ref{table:3d_shape_results}). Note that the weights for each trained MLP are only 2MB, while the triangle mesh files for the objects shown are 61MB, 7MB, 79MB, and 32MB respectively.

\subsection{2D CT} %%%%%%%%%%%%%%%%%%%%%%%

In computed tomography (CT), we observe measurements that are integral projections (integrals along parallel lines) of a density field.
We construct a 2D CT task by using ground truth $512\times512$ resolution images, and computing 20 synthetic integral projections at evenly-spaced angles. For each of these images, the supervision data is the set of integral projections, and the test PSNR is evaluated over the original image. 

We use two datasets for our 2D CT task: randomized Shepp-Logan phantoms~\cite{shepp}, and the ATLAS brain dataset~\cite{atlas}.
For each dataset, we perform a hyperparameter sweep over mapping scales on 8 examples. We found that scales $\sigma_g=4$ and $\sigma_p=3$ work best for the \emph{Shepp} dataset and $\sigma_g=5$ and $\sigma_p=5$ work best for the \emph{ATLAS} dataset. In Table~\ref{table:2d_ct_results}, we report model performance using the optimal mapping scale on a distinct set of 8 images.

\begin{table}[h!]
\centering
\begin{tabular}{l|cc}
            & Shepp & ATLAS \\ \hline
           No mapping      & $16.75 \pm 3.64$ & $15.44 \pm 1.28$ \\
           Basic           & $23.31 \pm 4.66$ & $16.95 \pm 0.72$ \\
           Positional enc. & $26.89 \pm 1.46$ & $19.55 \pm 1.09$ \\
           Gaussian        & $\mathbf{28.33 \pm 1.15}$ & $\mathbf{19.88 \pm 1.23 }$
\end{tabular}
\vspace{2mm}
\caption{2D CT results (mean $\pm$ standard deviation of PSNR).}
\label{table:2d_ct_results}
\end{table}

Each model (MLP with 4 layers, 256 channels, ReLU activation, sigmoid output) is trained for 1000 iterations using the Adam~\cite{adam} optimizer with default settings ($\beta_1=0.9$, $\beta_2=0.999$, $\epsilon=10^{-8}$). The learning rate is manually tuned for each method. Gaussian RFF and positional encoding use a learning rate of $10^{-3}$, and the basic and ``no mapping'' method use a learning rate of $10^{-2}$.

\subsection{3D MRI} %%%%%%%%%%%%%%%%%%%%%%%

In magnetic resonance imaging (MRI), we observe measurements that are Fourier coefficients of the atomic response to radio waves under a magnetic field.
We construct a toy 3D MRI task by using ground truth $96\times96\times96$ resolution volumes and randomly sampling $\sim\!13\%$ of the Fourier coefficients for each volume from an isotropic Gaussian. For each of these volumes, the supervision data is the set of sampled Fourier coefficients, and the test PSNR is evaluated over the original volume.

We use the ATLAS brain dataset~\cite{atlas} for our 3D MRI experiments.
We perform a hyperparameter sweep over mapping scales on 6 examples. We find that scales $\sigma_g=5$ and $\sigma_p=4$ perform best. In Table~\ref{table:3d_mri_results}, we report model performance using the optimal mapping scale on a distinct set of 6 images.
Each model (MLP with 4 layers, 256 channels, ReLU activation, sigmoid output) is trained for 1000 iterations using the Adam~\cite{adam} optimizer with default settings ($\beta_1=0.9$, $\beta_2=0.999$, $\epsilon=10^{-8}$). We use a manually-tuned learning rate of $2\times10^{-3}$ for each method. Results are visualized in Figure~\ref{fig:mri_atlas_supp}.

\begin{table}[h!]
\centering
\begin{tabular}{l|c}
            & ATLAS \\  \hline
           No mapping      & $26.14 \pm 1.45$\\
           Basic           & $28.58 \pm 2.45$\\
           Positional enc. & $32.23 \pm 3.08$\\
           Gaussian        & $\mathbf{34.51 \pm 2.72}$
\end{tabular}
\vspace{2mm}
\caption{3D MRI results (mean $\pm$ standard deviation of PSNR).}
\label{table:3d_mri_results}
\end{table}

\subsection{3D inverse rendering for view synthesis} %%%%%%%%%%%%%%%%%%%%%%%
In this task we use the ``tiny NeRF'' simplified version of the view synthesis method NeRF~\cite{mildenhall2020nerf} where hierarchical sampling and view dependence have been removed. The model is trained to predict the color and volume density at an input 3D point. Volumetric rendering is used to render novel viewpoints of the object. The loss is calculated between the rendered views and ground truth renders. In our experiments we use the NeRF \emph{Lego} dataset of 120 images downsampled to $400 \times 400$ pixel resolution. The dataset is split into 100 training images, 7 validation images, and 13 test images. The reconstruction quality on the validation images is used to determine the best mapping scale; for this scene we find $\sigma_g=6.05$ and $\sigma_p=1.27$ perform best.

The model (MLP with 4 layers, 256 channels, ReLU activation, sigmoid on RGB output) is trained for $5 \times 10^5$ iterations using the Adam~\cite{adam} optimizer with default settings ($\beta_1=0.9$, $\beta_2=0.999$, $\epsilon=10^{-8}$). The learning rate is manually tuned for each mapping: $10^{-2}$ for no mapping, $5 \times 10^{-3}$ for basic, $5 \times 10^{-4}$ for positional encoding, and $5 \times 10^{-4}$ for Gaussian. During training we use batches of 1024 rays.

The original NeRF method \cite{mildenhall2020nerf} uses an input mapping similar to the \emph{Positional encoding} we compare against. The original NeRF mapping is smaller than our mappings (8 vs. 256 frequencies). We include metrics for this mapping in Table~\ref{table:nerf_results} under \emph{Original pos. enc}. The positional encoding mappings only contain frequencies on the axes, and are therefore biased towards signals with on-axis frequency content (as demonstrated in Section~\ref{sec:axis-aligned}). In our experiments we rotate the \emph{Lego} scene, which was manually axis-aligned in the original dataset, for a more equitable comparison. Table~\ref{table:nerf_results} also reports metrics for positional encodings on the original axis-aligned scene. Results are visualized in Figure~\ref{fig:nerf_supp}.

\begin{table}[h!]
\centering
\begin{tabular}{l|c}
            & 3D NeRF \\  \hline
           No mapping      & $22.41 \pm 0.92$\\
           Basic           & $23.16 \pm 0.90$\\
           Original pos. enc. & $24.81 \pm 0.88$\\
           Positional enc. & $25.28 \pm 0.83$\\
           Gaussian        & $\mathbf{25.48 \pm 0.89}$\\ \hline
           Original pos. enc. (axis-aligned) & $25.60 \pm 0.76$\\
           Positional enc. (axis-aligned) & $26.27 \pm 0.91$\\
\end{tabular}
\vspace{2mm}
\caption{3D NeRF results (mean and standard deviation of PSNR). Error is calculated based on held-out images of the scene since the ground truth radiance field is not known.}
\label{table:nerf_results}
\end{table}

\clearpage
\section{Additional results figures}

\label{sec:supp_images}

\begin{figure}[p]
\centering

\begin{tabular}{@{}c@{\,\,}c@{\,\,}c@{\,\,}c@{\,\,}c@{}}
\includegraphics[width=\picwidth]{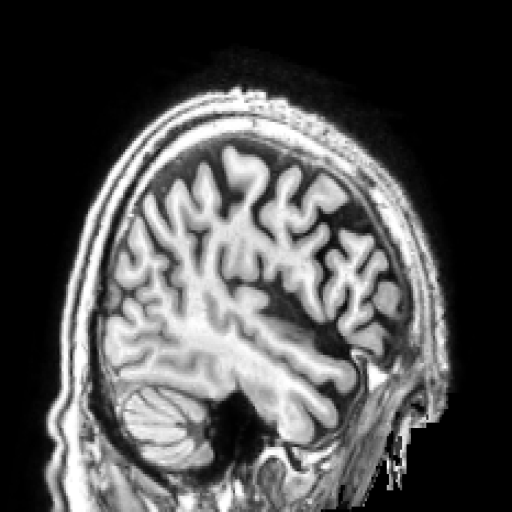} &
\includegraphics[width=\picwidth]{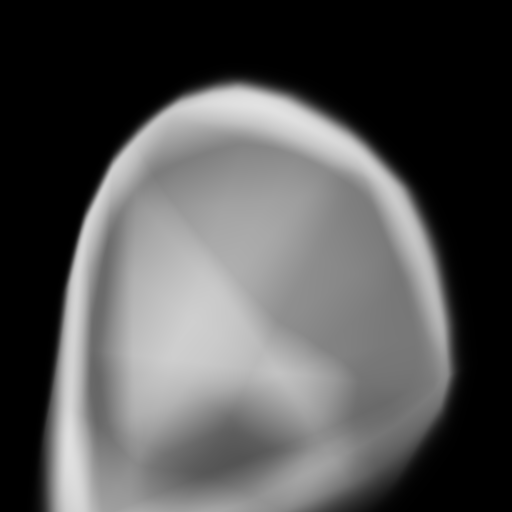} &
\includegraphics[width=\picwidth]{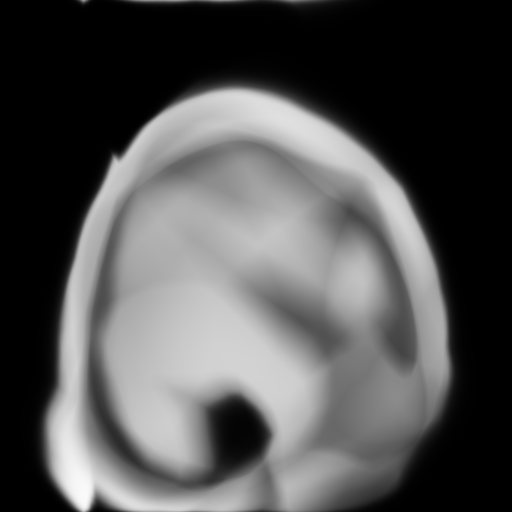} &
\includegraphics[width=\picwidth]{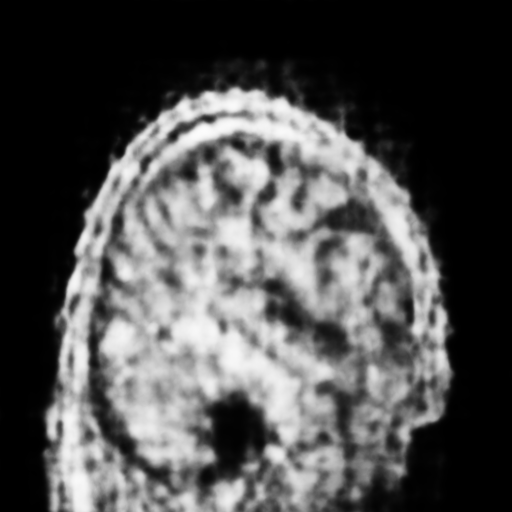} &
\includegraphics[width=\picwidth]{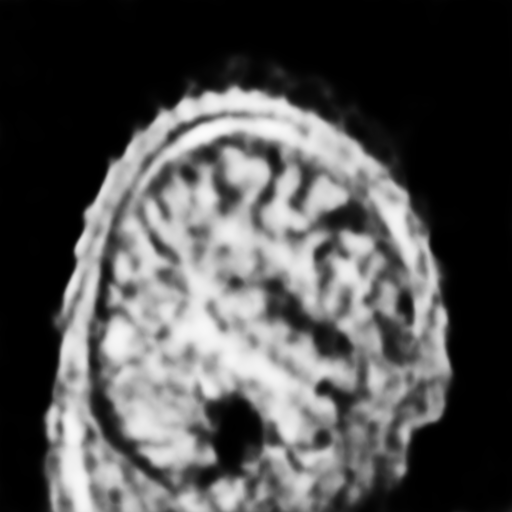}\\

\includegraphics[width=\picwidth]{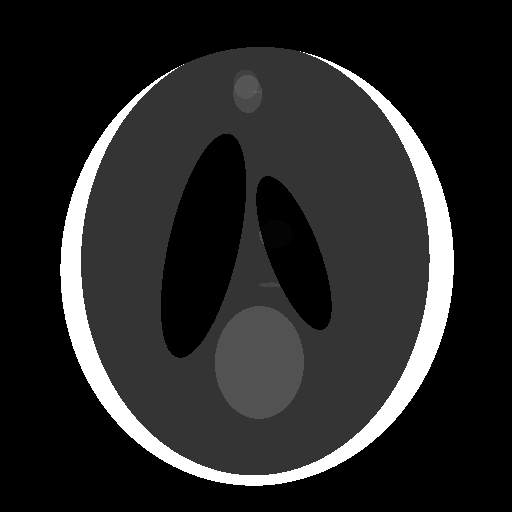} &
\includegraphics[width=\picwidth]{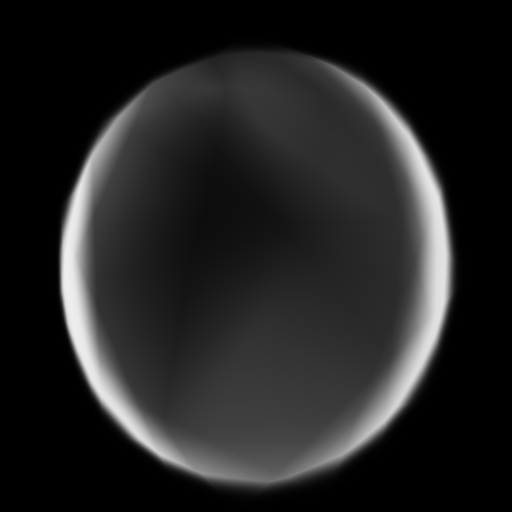} &
\includegraphics[width=\picwidth]{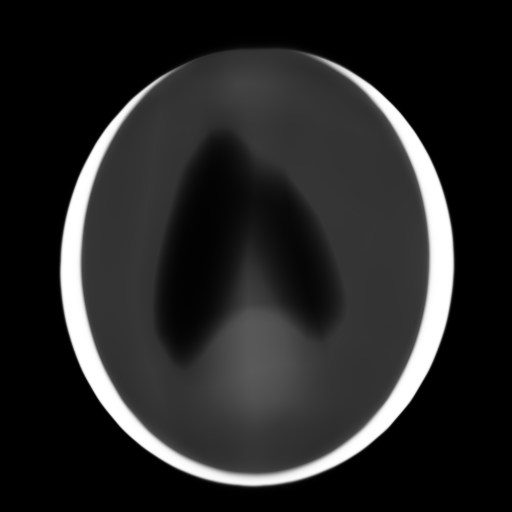} &
\includegraphics[width=\picwidth]{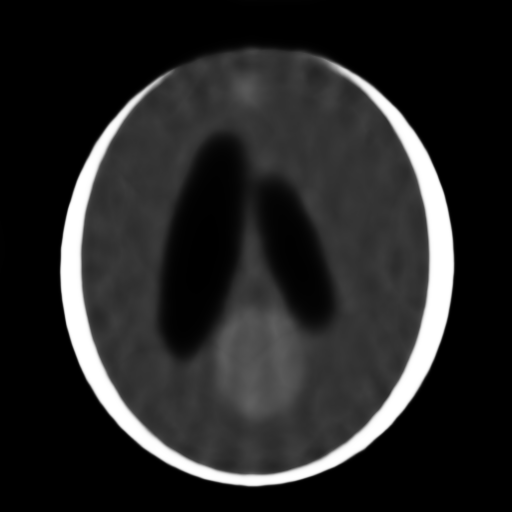} &
\includegraphics[width=\picwidth]{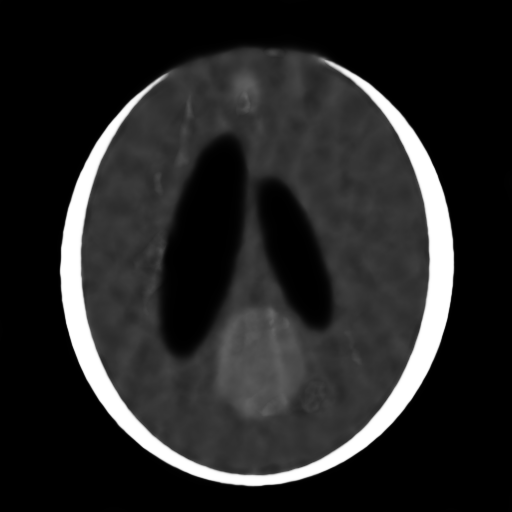}\\

\small{(a) Ground Truth}
& \small{(b) No mapping}
& \small{(c) Basic}
& \small{(d) Positional enc.}
& \small{(e) Gaussian}
\end{tabular}
\caption{Results for the 2D CT task.}
\label{fig:ct_atlas_supp}
\end{figure}

\begin{figure}[t]
\centering

\begin{tabular}{@{}c@{\,\,}c@{\,\,}c@{\,\,}c@{\,\,}c@{}}
\includegraphics[width=\picwidth]{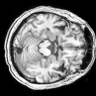} &
\includegraphics[width=\picwidth]{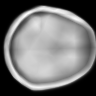} &
\includegraphics[width=\picwidth]{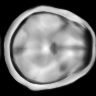} &
\includegraphics[width=\picwidth]{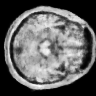} &
\includegraphics[width=\picwidth]{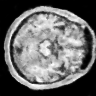}\\

\small{(a) Ground Truth}
& \small{(b) No mapping}
& \small{(c) Basic}
& \small{(d) Positional enc.}
& \small{(e) Gaussian}
\end{tabular}
\caption{Additional results for the 3D MRI task.}
\label{fig:mri_atlas_supp}
\end{figure}

\begin{figure}[p]
\centering

\begin{tabular}{@{}c@{\,\,}c@{\,\,}c@{\,\,}c@{\,\,}c@{}}
\includegraphics[width=\picwidth]{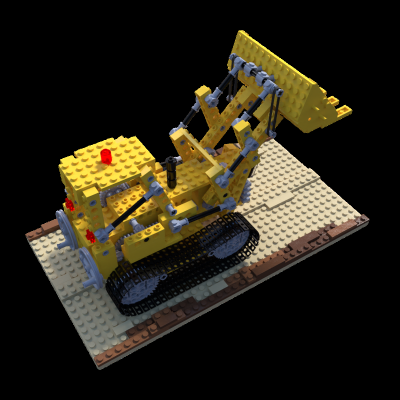} &
\includegraphics[width=\picwidth]{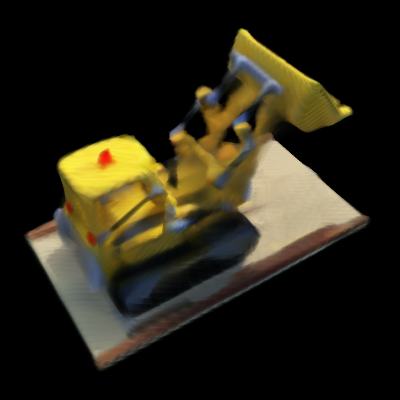} &
\includegraphics[width=\picwidth]{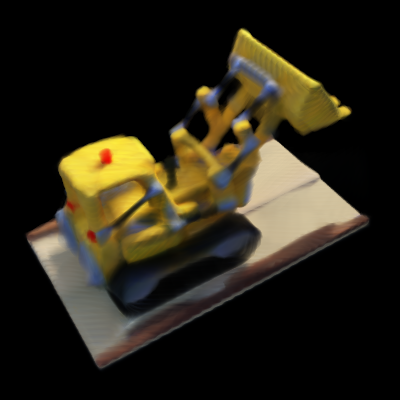} &
\includegraphics[width=\picwidth]{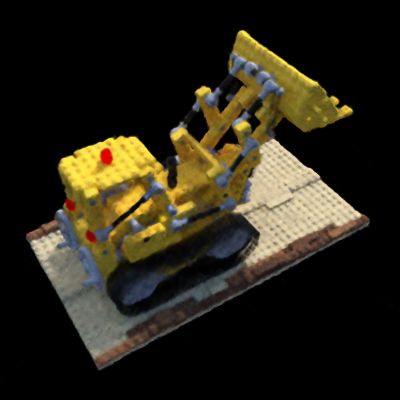} &
\includegraphics[width=\picwidth]{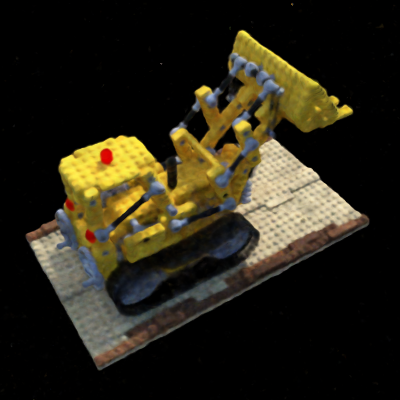}\\

\includegraphics[width=\picwidth]{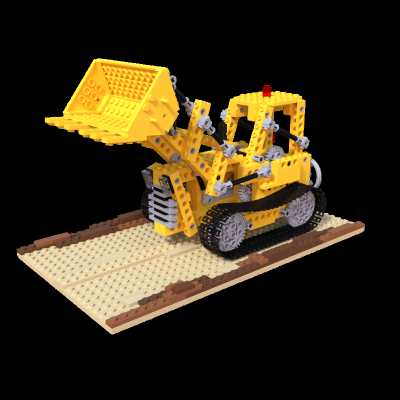} &
\includegraphics[width=\picwidth]{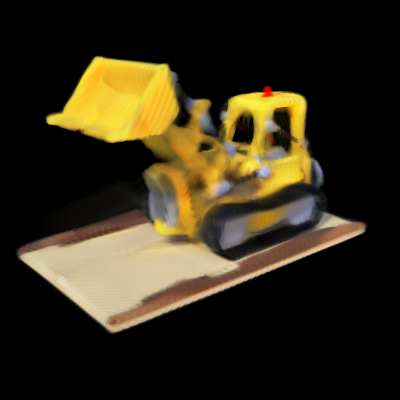} &
\includegraphics[width=\picwidth]{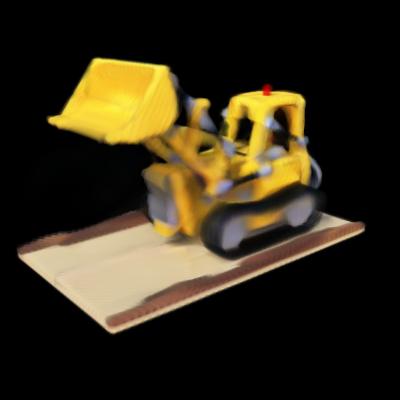} &
\includegraphics[width=\picwidth]{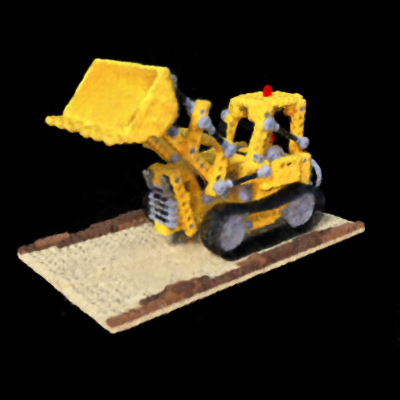} &
\includegraphics[width=\picwidth]{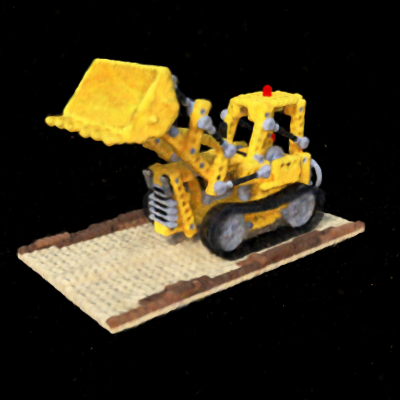}\\

\includegraphics[width=\picwidth]{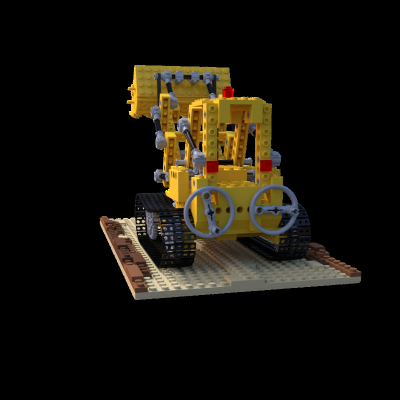} &
\includegraphics[width=\picwidth]{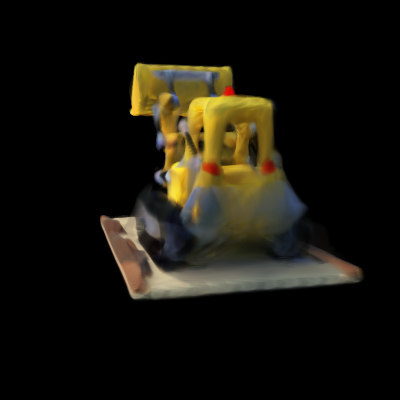} &
\includegraphics[width=\picwidth]{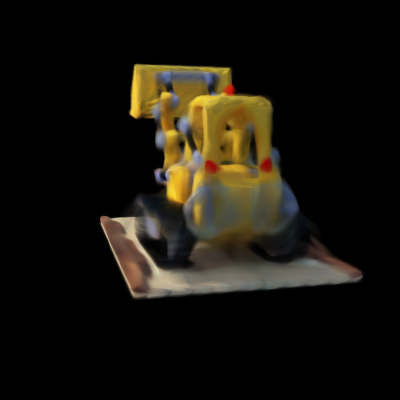} &
\includegraphics[width=\picwidth]{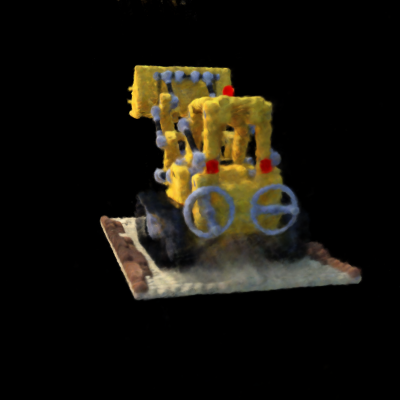} &
\includegraphics[width=\picwidth]{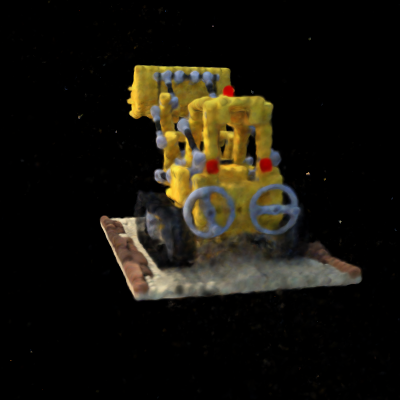}\\

\small{(a) Ground Truth}
& \small{(b) No mapping}
& \small{(c) Basic}
& \small{(d) Positional enc.}
& \small{(e) Gaussian}
\end{tabular}
\caption{Additional results for the inverse rendering task~\cite{mildenhall2020nerf}.}
\label{fig:nerf_supp}
\end{figure}

\end{appendices}

\end{document}